\documentclass[lettersize,journal]{IEEEtran}
\usepackage{amsmath,amsfonts}
\usepackage{algorithmic}
\usepackage{algorithm}
\usepackage{array}
\usepackage{textcomp}
\usepackage{stfloats}
\usepackage{url}
\usepackage{verbatim}
\usepackage{graphicx}
\usepackage{times}
\usepackage{epsfig}

\usepackage{bm}
\usepackage{booktabs}  
\renewcommand{\mathbf}{\boldsymbol}
\usepackage[dvipsnames]{xcolor}

\usepackage{amssymb}

\usepackage{multirow}
\usepackage{amsthm}
\usepackage{bm}
\usepackage{booktabs}  
\usepackage{pifont}
\usepackage{bbding}
\usepackage{subcaption}

\renewcommand{\mathbf}{\boldsymbol}

\def\x{\mathbf{x}}

\def\X{\mathbf{X}}
\def\C{\mathbf{C}}
\def\G{\mathbf{G}}

\def\P{\mathbf{P}}
\def\T{\mathbf{T}}

\def\F{\mathbf{F}}

\def\P{\mathbf{P}}

\newcommand{\ie}{\emph{i.e.}}
\newcommand{\eg}{\emph{e.g.}}

\usepackage{dsfont}

\begin{document}
	\newcolumntype{L}[1]{>{\raggedright\arraybackslash}p{#1}}
	\newcolumntype{C}[1]{>{\centering\arraybackslash}p{#1}}
	\newcolumntype{R}[1]{>{\raggedleft\arraybackslash}p{#1}}
	
	\title{Point-DAE: Denoising Autoencoders for Self-supervised Point Cloud Learning}
	
	\author{Yabin Zhang, Jiehong Lin, Ruihuang Li, Kui Jia, and Lei Zhang,~\IEEEmembership{Fellow,~IEEE}
		\thanks{Y. Zhang, R. Li, and L. Zhang are with the Department of Computing, The Hong Kong Polytechnic University, HongKong. E-mails:  csybzhang@comp.polyu.edu.hk, cslzhang@comp.polyu.edu.hk  Correspondence to: L. Zhang}
		\thanks{J. Lin and K. Jia are with the School of Electronic and Information Engineering, South China University of Technology, Guangzhou, China.}
	}
	
	\markboth{IEEE TRANSACTIONS ON NEURAL NETWORKS AND LEARNING SYSTEMS}%
	{Shell \MakeLowercase{\textit{et al.}}: A Sample Article Using IEEEtran.cls for IEEE Journals}
	
	
	\maketitle
	
	\begin{abstract}
		Masked autoencoder has demonstrated its effectiveness in self-supervised point cloud learning. Considering that masking is a kind of corruption, in this work we explore a more general denoising autoencoder for point cloud learning (Point-DAE) by investigating more types of corruptions beyond masking. Specifically, we degrade the point cloud with certain corruptions as input, and learn an encoder-decoder model to reconstruct the original point cloud from its corrupted version. 
		Three corruption families (\ie, density/masking, noise, and affine transformation) and a total of fourteen corruption types are investigated with traditional non-Transformer encoders.
		Besides the popular masking corruption, we identify another effective corruption family, \ie, affine transformation. The affine transformation disturbs all points globally, which is complementary to the masking corruption where some local regions are dropped. 
		We also validate the effectiveness of affine transformation corruption with the Transformer backbones, where we decompose the reconstruction of the complete point cloud into the reconstructions of detailed local patches and rough global shape, alleviating the position leakage problem in the reconstruction.
		Extensive experiments on tasks of object classification, few-shot learning, robustness testing, part segmentation, and 3D object detection validate the effectiveness of the proposed method.
		The codes are available at \url{https://github.com/YBZh/Point-DAE}.
	\end{abstract}
	
	
	
	\section{Introduction}
	\IEEEPARstart{L}{earning} effective feature representations from unlabeled data has been attracting growing attention since manual label annotation is labor-intensive and expensive, especially for 3D point cloud data.
	To this end, self-supervised learning (SSL) techniques have been proposed to generate supervision signals from unlabeled data themselves via carefully designed pretext tasks, such as jigsaw puzzles \cite{noroozi2016unsupervised,sauder2019self}, instance contrastive discrimination \cite{chen2020simple,he2020momentum,sanghi2020info3d}, and masked autoencoder (MAE) \cite{he2021masked,yu2021point,pang2022masked}.
	Among those pretext tasks, the effectiveness of MAE has been validated in many applications \cite{he2021masked,pang2022masked,tong2022videomae}, including point cloud understanding \cite{wang2021unsupervised,yu2021point,pang2022masked}. From the input point clouds masked at a high ratio, an encoder-decoder model is trained to reconstruct the complete point clouds \cite{wang2021unsupervised,yu2021point,pang2022masked}. The pre-trained encoder is then applied to various downstream tasks, while the decoder is typically discarded. Different masking strategies, backbones, and reconstruction targets have been investigated under the MAE framework \cite{wang2021unsupervised,pang2022masked,zhang2022point,liu2022masked,zhang2022masked}.

\begin{figure*}
		\centering
		\includegraphics[width=0.93\linewidth]{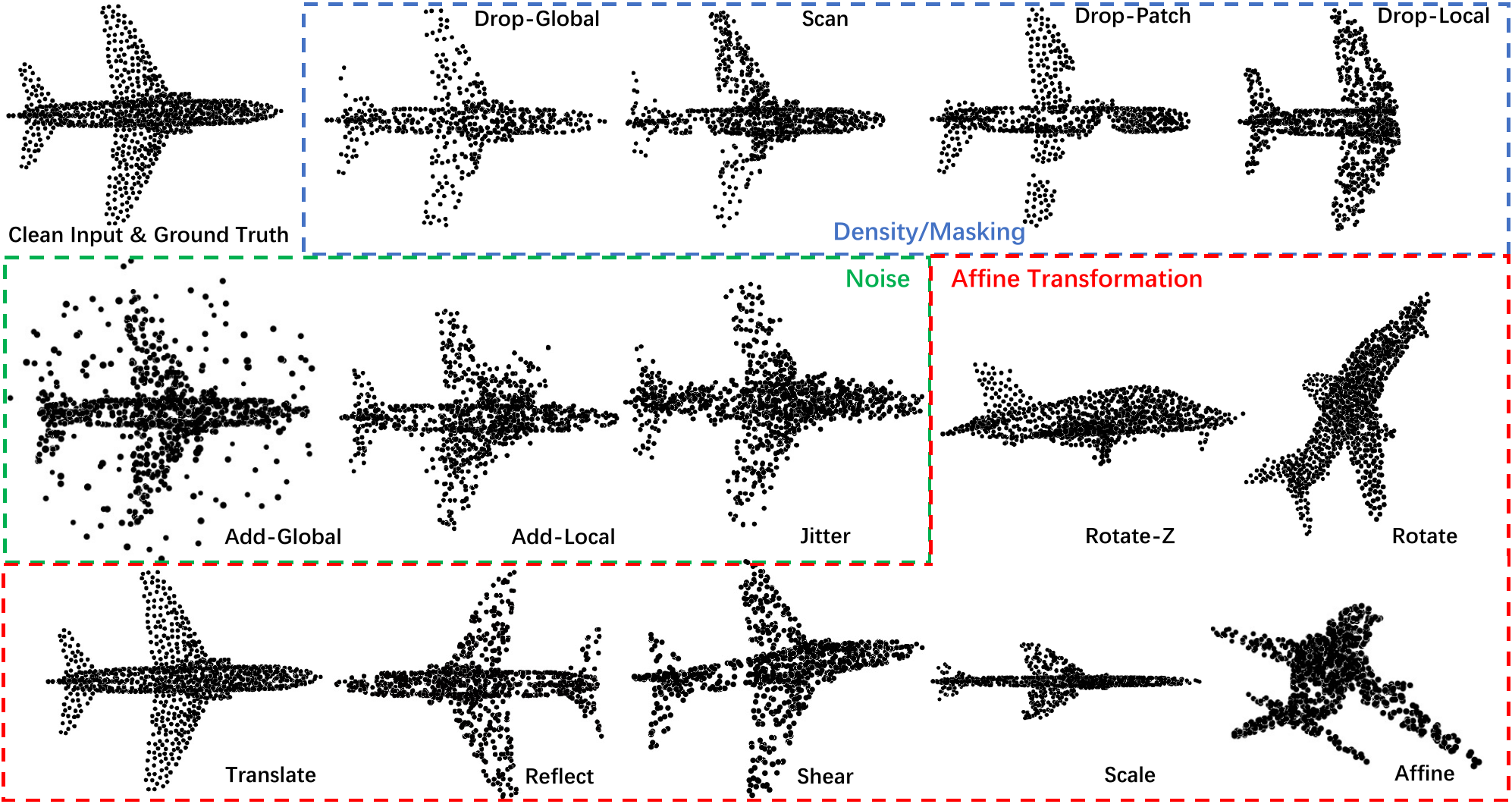}
		\vspace{-0.1cm}
		\caption{Illustration of the $14$ corruptions studied in this work, which can be classified into three corruption families, \ie, density/masking, noise, and affine transformation. 
			Please refer to the \textbf{Supplementary Material} for more detailed implementation of these corruptions.
			\vspace{-0.2cm}
		}\label{Fig:corruptions_vis}
	\end{figure*}

	Though MAE and its variants have achieved great successes, paying attention only to the masking strategy may limit the capacity of such a generation-based SSL framework. Considering that masking is a kind of corruption, we propose to explore a more general denoising autoencoder \cite{vincent2008extracting} for point cloud learning, namely Point-DAE, by investigating more types of corruptions beyond masking. Specifically, we first degrade the point cloud input with certain corruptions, and then learn an encoder-decoder model to reconstruct the uncorrupted input from its corrupted counterpart. 
	We investigate three corruption families, including density/masking, noise, and affine transformation, and hence a total of $14$ corruption types, as visualized in Fig. \ref{Fig:corruptions_vis}.  
	In other words, we study $14$ pretext tasks since each corruption instantiates a unique task.
	
	\begin{figure}
		\centering\includegraphics[width=.99\linewidth]{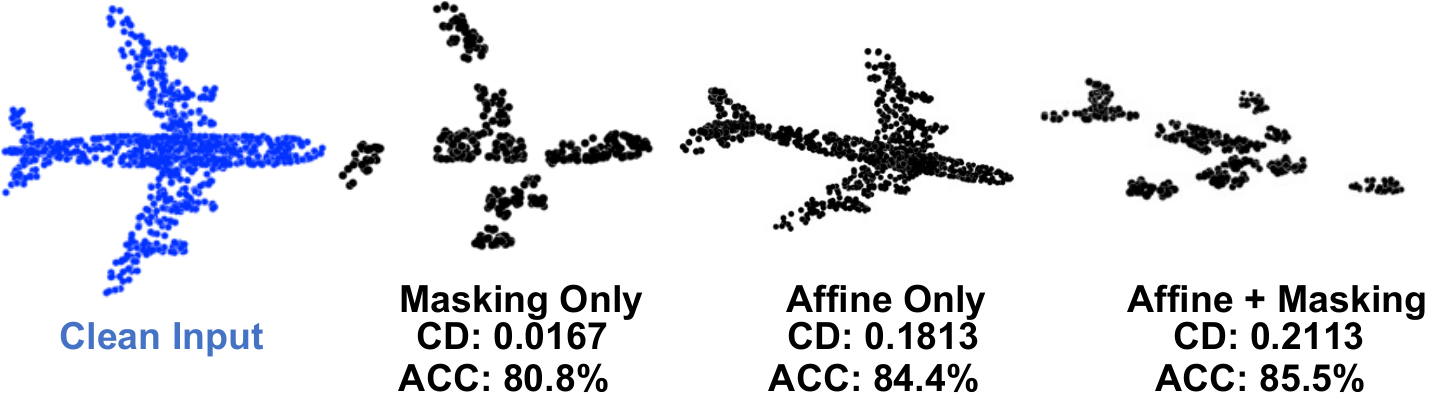}
		\caption{Visualizations of masking (\eg, Drop-Local) and affine transformation (\eg, Affine) corruptions and the corresponding Chamfer Distance (CD) to the clean input, where the reported CD values are averaged over the ShapeNet training set. The ACC reports the classification accuracy on downstream ScanObjectNN dataset, as detailed in Tab. \ref{Tab:results_corruptions}.} \label{Fig:mask_affine_cdloss}
		\vspace{-0.4cm}
	\end{figure}
	
	\begin{figure*}[h!]
	\centering
\[
\begin{array}{c}
\parbox{15cm}{
	\[
	\left[
	\begin{array}{cccc}
	\cos(\theta_z) & -\sin(\theta_z) & 0 & 0 \\
	\sin(\theta_z) & \cos(\theta_z) & 0 & 0 \\
	0 & 0 & 1 & 0 \\
	0 & 0 & 0 & 1 \\
	\end{array}
	\right]
	\left[
	\begin{array}{cccc}
	\cos(\theta_y) & 0 & \sin(\theta_y) & 0 \\
	0 & 1 & 0 & 0 \\
	-\sin(\theta_y) & 0 & \cos(\theta_y) & 0 \\
	0 & 0 & 0 & 1 \\
	\end{array}
	\right]
	\left[
	\begin{array}{cccc}
	1 & 0 & 0 & 0 \\
	0 & \cos(\theta_x) & -\sin(\theta_x) & 0 \\
	0 & \sin(\theta_x) & \cos(\theta_x) & 0 \\
	0 & 0 & 0 & 1 \\
	\end{array}
	\right]
	\]
	\begin{center}
		\text{Rotate}
	\end{center}
}
\\
\begin{array}{ccc}
	\parbox{4cm}{
		\[
		\left[
		\begin{array}{cccc}
		1 & 0 & 0 & \tau_x \\
		0 & 1 & 0 & \tau_y \\
		0 & 0 & 1 & \tau_z \\
		0 & 0 & 0 & 1 \\
		\end{array}
		\right]
		\]
		\begin{center}
			\text{Translate}
		\end{center}
	}
	\parbox{4cm}{
		\[
		\left[
		\begin{array}{cccc}
		r_x & 0 & 0 & 0 \\
		0 & r_y & 0 & 0 \\
		0 & 0 & r_z & 0 \\
		0 & 0 & 0 & 1 \\
		\end{array}
		\right]
		\]
		\begin{center}
			\text{Reflect}
		\end{center}
	}
\parbox{4cm}{
	\[
	\left[
	\begin{array}{cccc}
	1 & s_{xy} & s_{xz} & 0 \\
	s_{yx} & 1 & s_{yz} & 0 \\
	s_{zx} & s_{zy} & 1 & 0 \\
	0 & 0 & 0 & 1 \\
	\end{array}
	\right]
	\]
	\begin{center}
		\text{Shear}
	\end{center}
}
\parbox{4cm}{
	\[
	\left[
	\begin{array}{cccc}
	s_x & 0 & 0 & 0 \\
	0 & s_y & 0 & 0 \\
	0 & 0 & s_z & 0 \\
	0 & 0 & 0 & 1 \\
	\end{array}
	\right]
	\]
	\begin{center}
		\text{Scale}
	\end{center}
}
\end{array}
\end{array}
\]
    \caption{\textcolor{black}{Transformation matrices corresponding to different sub-transformations of the affine transformation family. The parameters $\theta_x, \theta_y, \theta_z  \in [-\pi, \pi]$ respectively determine the rotation angles around the X,Y, and Z axes in the Rotate corruption. 
    Rotate-Z is a subset of the Rotate transformation, where the point cloud is rotated  around the Z axis only, with $\theta_x, \theta_y=0$ and $\theta_z  \in [-\pi, \pi]$.
    The parameters $\tau_x, \tau_y, \tau_z \in \mathbb{R}$ respectively decide the translate magnitude along the X,Y, and Z axes in the Translate corruption.
    For the Reflect transformation, we set $r_x, r_y, r_z \in \{ -1, 1\}$, and reflect the point cloud along the X/Y/Z axis with $r_x/r_y/r_z = -1$. The parameters $s_{xy}, s_{xz}, s_{yx}, s_{yz}, s_{zx}, s_{zy} \in \mathbb{R}$ determine the shear magnitude and $s_x, s_y, s_z \in \mathbb{R}^+$ decide the scale intensity in Shear and Scale corruptions, respectively.
    The full Affine corruption is obtained by combining these sub-transformations, with its transformation matrix resulting from the multiplication of the matrices of these sub-ones.
    We conduct detailed analyses on these hyper-parameters in the Supplementary Material. }}
\label{fig:transformation_matrices}
\end{figure*}

	By analyzing these pretext tasks, we identify another effective corruption family, \ie, affine transformation, in addition to the popular masking strategy.
	With affine transformation, we disturb all points with a specific transformation matrix globally. Such a global transformation introduces considerable shape deformation, as revealed by the enlarged CD in Fig. \ref{Fig:mask_affine_cdloss}. Considering that the masking corruption drops partial points of local regions while the affine transformation  introduced global distortion, better results are expected by combining these two complementary corruption types, as justified in Tab. \ref{Tab:results_corruptions}. 
	
	While it is direct to apply the affine transformation to the traditional non-Transformer backbones, introducing it to the Transformer backbones is non-trivial. Specifically, existing Transformer-based methods \cite{yu2021point,pang2022masked} typically adopt unordered point patches as input and reconstruct the local point patches guided by their global position, \eg, the patch centers via positional embedding. Such leakage of global position impedes the perception of global shape during pre-training. This is even more serious with the affine transformation corruption, where the deformed global shape is directly indicated by the leaked global position, diluting the influence of affine transformation.
	To tackle this problem, we propose a simple approach to decompose the reconstruction of the complete point cloud into the reconstructions of detailed local patches and rough global shape. By modeling the global shape explicitly, the effectiveness of global affine transformation is fully utilized and improved model representation can be achieved, as presented in Tab. \ref{Tab:global_local_losses}.
	
	We carefully analyze the proposed components and validate the effectiveness of our Point-DAE on a wide range of downstream tasks.
	Specifically, we testify the complementarity between corruptions of masking and affine transformation on both Transformer and non-Transformer encoders.
	With frozen encoders, Point-DAE achieves 87.9\% and 84.6\% classification accuracy on ScanObjectNN dataset with DGCNN and Transformer encoders, outperforming the second-best methods \cite{afham2022crosspoint,pang2022masked} by 6.2\% and 5.9\%, respectively.
	When fine-tuning on downstream tasks, Point-DAE achieves 92.1\% (+8.2\%) accuracy on ScanObjectNN for shape classification, 86.0\% (+0.5\%) instance mIoU on ShapeNetPart for part segmentation, 93.0\% (+9.4\%) few-shot classification accuracy on 10-way 10-shot on ModelNet40, and 0.835 mCE (-0.165 mCE) on ModelNet-C for robustness testing with the DGCNN encoder. Similar improvements are achieved with the Transformer backbone, validating the generality of our method. We summarize our contributions as follows:
	\begin{itemize}
		\item We investigate a general denoising autoencoder framework for self-supervised point cloud learning (Point-DAE) by exploring three corruption families (\eg, density/masking, noise, and affine transformation) and a total of fourteen corruption types. By analyzing these variants, we identify another effective corruption family, \ie, affine transformation, which is complementary to the popular masking strategy.
		\item We validate the effectiveness of the revealed affine transformation corruption on various backbone encoders, including the non-Transformer and Transformer ones. For the Transformer-based Point-DAE, we propose a reconstruction decomposition strategy to tackle the position leakage problem occurred in the reconstruction.
		\item We carefully ablate the proposed components and validate the pre-trained models with extensive downstream tasks, including object classification, few-shot learning, robustness testing, part segmentation, and 3D object detection. We also present intuitive reconstruction visualization and some interesting observations, such as the influence of object poses in the pre-training dataset.
	\end{itemize}

	\section{Related Work}
	
	\textbf{Pre-training by Reconstruction.}
	Feature learning via sample reconstruction has a long research history \cite{hinton2006reducing,vincent2008extracting}. Recently, masked modeling has gained increasing attention in multiple data modalities, where we mask some parts of the data signals as input and learn an encoder-decoder model to reconstruct the masked data.
	The pioneering masked language modeling has achieved a significant performance boost on downstream tasks \cite{radford2018improving,devlin2018bert}.  Inspired by this, masked image modeling \cite{bao2021beit,he2021masked} and masked video modeling \cite{tong2022videomae,feichtenhofer2022masked} are actively investigated and present considerable successes.
	Recently, masked modeling is also explored in the point cloud domain with irregular data input.
	An initial attempt masks points via view occlusion \cite{wang2021unsupervised}, and recent works mask point patches with the Transformer backbone \cite{vaswani2017attention}.
	Given discrete tokens from point patches via dVAE \cite{rolfe2016discrete}, Point-BERT \cite{yu2021point} reconstructs masked tokens with a contrastive objective. Without dVAE mapping, Point-MAE \cite{pang2022masked} and Point-M2AE \cite{zhang2022point} reconstruct the masked point patches directly with a pure masked autoencoding framework. 
	Besides explicit points, the target of implicit 3D representations, such as occupancy value and signed distance function, is also studied given masked point cloud \cite{liu2022masked,yan2022implicit}.
	Unlike these masked modeling methods that only corrupt points by masking local regions, we investigate a more general denoising autoencoder by conducting sample corruption in a broader space. With such a comprehensive study, we recommend to use another global affine transformation corruption to complement the local masking strategy, leading to more effective and robust feature representation for point clouds.

	\textbf{Self-supervised Point Cloud Learning.}
	The performance of point cloud learning has been boosted with carefully designed model architectures and optimization strategies under the supervised learning context \cite{li2018pointcnn,thomas2019kpconv,wu2019pointconv,qiu2021dense,zhao2021point,ma2022rethinking,liu2020closer,wu2022point,xu2021paconv,qian2022pointnext,guo2021pct,xiang2021walk}. In this work, we explore an orthogonal SSL direction to learn effective feature representations from unlabeled point cloud data. 
	Pretext tasks such as Jigsaw \cite{sauder2019self} and orientation estimation \cite{poursaeed2020self} are explored in pioneering attempts. 
	One mainstream pipeline is contrastive learning, which minimizes the distance between different `views' of the same sample and maximizes the distance across different samples. The required sample view can be introduced via different modalities \cite{afham2022crosspoint}, data augmentation \cite{sanghi2020info3d,zhang2021self}, view occlusion \cite{xie2020pointcontrast}, and local structure \cite{rao2020global}.
	Some recent methods \cite{zhang2023learning,qi2023contrast} require additional knowledge from pretrained models of other modalities, resulting in unfair advantages.
	Reconstruction is another popular SSL framework, which reconstructs the clean point cloud from its corrupted version \cite{chen2021shape,pang2022masked,zhang2022masked}. The corrupted point cloud can be generated by part disorganization \cite{chen2021shape}, contour perturbation \cite{xu2022cp}, and masking \cite{wang2021unsupervised,yu2021point,pang2022masked,zhang2022point}. 
	Most of these corruptions only corrupt partial points of the input, while other points remain unchanged.
	In contrast, we promote using affine transformation to corrupt all input points globally. 
	We validate the complementary between the popular masking and the studied affine transformation on various encoders and achieve state-of-the-art performance on wide downstream tasks.
	
	\textbf{Sample corruptions.}
	Sample corruptions have been widely investigated in different research areas. 
	Data augmentation via sample corruptions has demonstrated its effectiveness in settings of both independent and identical distribution \cite{cubuk2019autoaugment,cubuk2020randaugment} and our-of-distribution (OOD) \cite{hendrycks2019augmix,zhang2022exact,zhang2023adversarial,zhang2020unsupervised}. The dataset for testing OOD generalization performance is typically generated by applying corruptions to clean samples \cite{hendrycks2019benchmarking,ren2022modelnetc}.
	Sample corruptions also play important roles in semi-supervised learning \cite{sohn2020fixmatch} and contrastive learning \cite{chen2020simple}. 
	In low-level vision tasks (\eg, image super-resolution \cite{dong2015image} and point cloud upsampling \cite{yu2018pu}), training pairs are typically constructed by applying corruptions to high-quality samples. However, the objective in low-level vision tasks is mainly to reconstruct high-quality samples, while the main objective of SSL is to learn effective feature representations without manual annotations.
	
	
	\section{Our Approach} \label{Sec:methods}
	We first present an overview of the proposed Point-DAE method. Then we present how to implement it with non-Transformer and Transformer encoders, respectively.
	
	\subsection{Overview of Point-DAE}
\begin{figure*}
		\centering
		\includegraphics[width=0.94\linewidth]{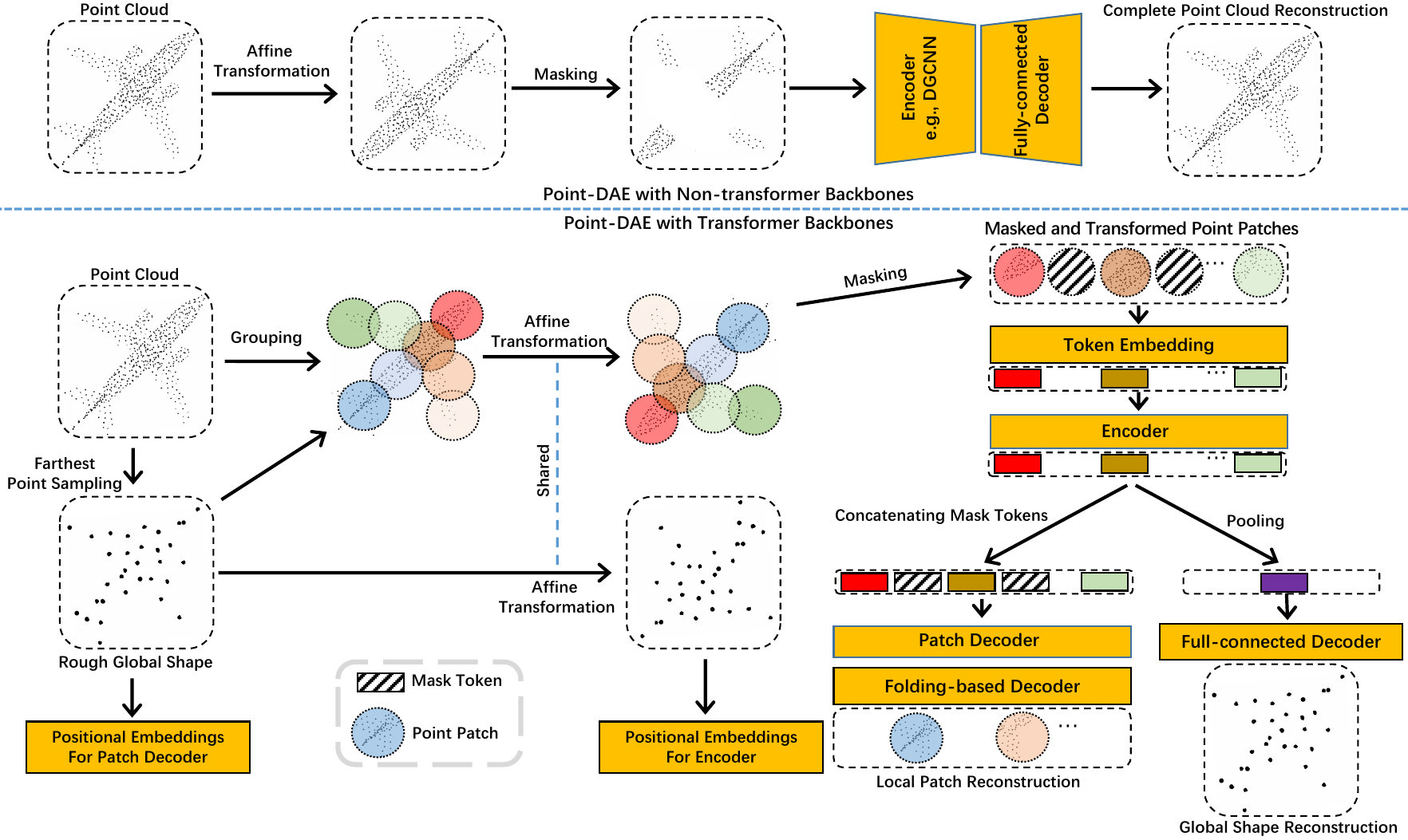}
		\caption{An overview of Point-DAE with non-Transformer backbones (upper part) and Transformer backbones (lower part), where we use a toy rotation operation to represent the affine transformation for visualization.
		}\label{Fig:methods} 
	\end{figure*}

	The Point-DAE method can be summarized as follows:
	\begin{equation} \label{Equ:pretrain}
	\min_{\mathrm{E},\mathrm{D}} \mathbb{E}_{\X\in \mathbb{X}} \mathcal{L}\left(\mathrm{D}(\mathrm{E}(\mathrm{Corruption}(\X))), \X\right),
	\end{equation}
	where $\X \in \mathbb{R}^{w\times 3}$ is a point cloud in the dataset $\mathbb{X}$. $\mathrm{E}(\cdot)$ and $\mathrm{D}(\cdot)$ represent the encoder and decoder models, respectively. $\mathrm{Corruption}(\cdot)$ indicates the applied corruption (\eg, masking and affine transformation). $\mathcal{L}(\cdot,\cdot)$ is the loss function measuring the reconstruction quality, which is set as the popular CD loss \cite{fan2017point} by default: 
	\begin{equation} \label{Equ:cd}
	\mathcal{L}_{cd}(\widehat{\X}, \X) = \frac{1}{w} \sum_{\x \in \X} \min_{\widehat{\x} \in \widehat{\X}} \| \x - \widehat{\x} \|_2^2 + \frac{1}{\widehat{w}} \sum_{\widehat{\x} \in \widehat{\X}} \min_{\x \in \X} \| \x - \widehat{\x} \|_2^2,
	\end{equation}
	where $\widehat{\X} \in \mathbb{R}^{\widehat{w}\times 3}$ is the reconstructed point cloud. 
	$\x \in \mathbb{R}^3$ and $\widehat{\x} \in \mathbb{R}^3$ are points in $\X$ and $\widehat{\X}$, respectively.
	We explore Point-DAE on various encoder backbones, which can be roughly clustered into two clans, \ie, the non-Transformer clan that processes the global point cloud as a whole \cite{qi2017pointnet,qi2017pointnet++,wang2019dynamic} and the Transformer clan that divides the whole point cloud into local patches as input \cite{yu2021point,pang2022masked,zhang2022point}.
	The corresponding implementations are illustrated in Fig. \ref{Fig:methods} and detailed in the following sections.
	
	The corruption $\mathrm{Corruption}(\cdot)$ is our focus in this work, given that each type of corruption instantiates a unique pretext task.
	Inspired by the practice of \cite{sun2022benchmarking,ren2022modelnetc}, we systemically investigate three corruption families (\ie, density/masking, noise, and affine transformation) and a total of $14$ corruption types.
	Generally speaking, we randomly change the point density either uniformly or nonuniformly to implement the density/masking corruptions, and add random noises to the point cloud either locally or globally to implement the noise corruptions. The affine transformation corruptions are implemented by multiplying the augmented input with a random affine transformation matrix as follows:
	\begin{equation} \label{Equ:affine}
	\left[
	\begin{array}{c}
	x' \\
	y' \\
	z' \\
	1 \\
	\end{array}
	\right] =
	\left[
	\begin{array}{cccc}
	a_{11} & a_{12} & a_{13} & a_{14} \\
	a_{21} & a_{22} & a_{23} & a_{24} \\
	a_{31} & a_{32} & a_{33} & a_{34} \\
	0 & 0 & 0 & 1 \\
	\end{array}
	\right]
	\left[
	\begin{array}{c}
	x \\
	y \\
	z \\
	1 \\
	\end{array}
	\right],
	\end{equation}
	where $[x, y, z]$ and $[x',y', z']$ are the input and transformed points, respectively. The affine transformation matrix is defined by the $12$ parameters, \ie, $a_{ij}$, covering the Rotate, Translate, Reflect, Shear, and Scale sub-transformations with various magnitudes.
    \textcolor{black}{The matrices corresponding to these sub-transformations are illustrated in Fig. \ref{fig:transformation_matrices}.}

	
	
	The representative corruptions are visualized in Fig. \ref{Fig:corruptions_vis}. One can see that we implement four corruptions for density/masking family, three corruptions for noise family, and seven corruptions for affine transformation family, resulting in 14 types of corruptions in total. 
	Although we practically investigate all the 14 corruption types in Sec. \ref{Sec:experiments}, we only introduce the detailed implementation of the most effective variant, \ie, Point-DAE with affine transformation and masking corruptions, on different encoder backbones in the following sections.

	\subsection{Point-DAE with Non-Transformer Encoders} \label{Subsec:non_transformer}
	
	\textbf{Training Data Preparation.}
	Given a point cloud $\X \in \mathbb{R}^{w\times 3}$, we first generate its affine transformed counterpart $\X' \in \mathbb{R}^{w\times 3}$ by applying the random affine transformation matrix of Eq. (\ref{Equ:affine}) to $\X$ in a point-wise manner.
	
	Then, we mask some points from $\X'$. Existing approaches typically mask points occluded in a camera view \cite{wang2021unsupervised} or drop fixed-sized KNN clusters \cite{pang2022masked}. Unlike these methods, we promote a more effective strategy, \ie, Drop-Local,  by masking several random-sized KNN clusters. Specifically, we first randomly set the number of dropped clusters $\kappa \in \mathbb{Z}^+$ and the total number of dropped points $\eta = \lfloor \alpha w \rfloor$, where $\alpha \in (0,1)$ is the masking ratio. Then, for the $i$-th cluster, we randomly select a point in $\X'$ as the cluster center and drop its nearest $\eta_i$ points, where $\sum_{i=1}^{\kappa}\eta_i = \eta$ and $\eta_i > 0$. Finally, we get the masked and affine transformed $\X'_{mask}$.

	\textbf{Encoder.} 
	The encoder takes the $\X'_{mask}$ as input and outputs a global feature $\F'_e \in \mathbb{R}^d$:
	\begin{equation}
	\F'_e = \mathrm{E}(\X'_{mask}).
	\end{equation}
	The encoder $\mathrm{E}(\cdot)$ can be instantiated as any point cloud encoders \cite{qi2017pointnet,qi2017pointnet++,wang2019dynamic}, including the recent Transformer-based ones \cite{yu2021point,zhang2022point,zhao2021point}.  However, we promote a more effective implementation of Point-DAE for the Transformer backbones, as detailed in Sec. \ref{Subsec:transformer_methods}.

	\textbf{Decoder.} Given the encoded feature $\F'_e$, an decoder is learned to reconstruct the original point cloud $\X$. Inspired by the practice of point cloud reconstruction, we investigate two types of decoders, \ie, the fully-connected decoder $\mathrm{D}_{fc}$ \cite{achlioptas2018learning,yuan2018pcn} and the folding-based decoder $\mathrm{D}_{fold}$ \cite{yang2018foldingnet}.
	Specifically, the $\mathrm{D}_{fc}$ transforms $\F'_e$ into a vector of $3w$ elements with fully connected (FC) layers and reshapes it to reconstruct the point cloud $\widehat{\X} \in \mathbb{R}^{w\times 3}$:
	\begin{equation} \label{Equ:fc_decoder}
	\widehat{\X} = \mathrm{D}_{fc}(\F'_e) = \mathrm{Reshape}\left( FC (\F'_e) \right).
	\end{equation}
	The folding-based decoder $\mathrm{D}_{fold}$ \cite{yang2018foldingnet} generates the point cloud by deforming a canonical 2D grid $\G \in \mathbb{R}^{w\times 2}$ conditioned on the global feature $\F'_e$:
	\begin{equation} \label{Equ:folding}
	\widehat{\X} = \mathrm{D}_{fold} (\F'_e) = \mathrm{MLP}(\mathrm{Concat}[\G, \F'_e]),
	\end{equation}
	where $\mathrm{Concat}[\G, \F'_e] \in \mathbb{R}^{w\times (d+2)}$ concatenates the 2D grid $\G$ and the replicated global feature $\F'_e$, and $\mathrm{MLP}(\cdot):  \mathbb{R}^{w\times (d+2)} \to  \mathbb{R}^{w\times 3}$ constructs the point cloud with multi-layer perceptrons (MLP). 
	

	The decoder architecture also plays an important role in SSL, which is less studied.
	We find that the fully-connected decoder $\mathrm{D}_{fc}$ performs better in complete/global shape reconstruction, while the folding-based decoder $\mathrm{D}_{fold}$ shows advantages on modeling smooth local patches (please refer to Sec. \ref{Subsec:transformer_methods} and Sec. \ref{Sec:experiments} for more details).

	\textbf{Final Loss.} 
	We adopt the CD loss in Eq. (\ref{Equ:cd}) to measure the reconstruction divergence between $\X$ and $\widehat{\X}$, resulting in the following loss function:
	\begin{equation} \label{Equ:loss_final}
	\mathcal{L}_{all} =  \mathcal{L}_{cd}(\widehat{\X}, \X).
	\end{equation}

	\subsection{Point-DAE with Transformer Encoders} \label{Subsec:transformer_methods}
	Instead of reconstructing the complete point cloud directly as in Sec. \ref{Subsec:non_transformer}, we decompose the overall reconstruction task into the reconstructions of detailed local patches and rough global shape for Transformer backbones, which is detailed in the following content.
	
	\textbf{Training Data Preparation.}
	Given a point cloud $\X \in \mathbb{R}^{w\times 3}$, we first sample $n$ points as patch centers $\C \in \mathbb{R}^{n\times 3}$ via the Farthest Point Sampling method \cite{qi2017pointnet++}. Then the point patches $\P \in \mathbb{R}^{n \times k \times 3}$ are formed by searching the $K$ nearest points around each patch center using the KNN method.
	Then, we apply the masking and affine transformation corruptions to the point patches $\P$ in a patch-wise manner. 
	Specifically, we first generate the affine transformed patches $\P' \in \mathbb{R}^{n \times k \times 3}$ and patch centers $\C' \in \mathbb{R}^{n\times 3}$ by applying certain affine transformation matrix (cf. Eq. (\ref{Equ:affine})) to $\P$ and $\C$ in a point-wise manner. 
	Then, we normalize each point patch in $\P'$ and $\P$ by subtracting its patch center from the point coordinates.
	
	
	Given the point patches $\P'$, we randomly mask a large portion of them to conduct the masking corruption. Specifically, we mask out $m= \lfloor \alpha n \rfloor$ patches to obtain the masked patches  $\mathbf{P}'_{mask} \in \mathbb{R}^{m \times k \times 3}$ and unmasked patches $\mathbf{P}'_{vis} \in \mathbb{R}^{(n-m)\times k \times 3}$, where $\alpha \in (0,1)$ is the masking ratio.
	The corresponding masked patch centers $\C'_{mask} \in \mathbb{R}^{m\times 3}$ and unmasked patch centers $\C'_{vis} \in \mathbb{R}^{(n-m)\times 3}$ are similarly obtained for the usage of positional embedding (PE). 
	With the same masking strategy, the corresponding masked patches  $\P_{mask} \in \mathbb{R}^{m \times k \times 3}$, unmasked patches $\P_{vis} \in \mathbb{R}^{(n-m)\times k \times 3}$, masked patch centers $\C_{mask} \in \mathbb{R}^{m\times 3}$, and unmasked patch centers $\C_{vis} \in \mathbb{R}^{(n-m)\times 3}$ from $\P$ and $\C$ are also generated.
	
	The transformed and unmasked patches $\P'_{vis}$ and their centers $\C'_{vis}$ are used as the input to the Transformer-based encoder. The vanilla patch centers $\C$ are adopted as the supervision of global shape reconstruction and the PE for the Transformer-based patch decoder. The vanilla masked patches $\P_{mask}$ are adopted as the supervision signal for local patches reconstruction. The details are presented in the following subsections.
	
	\textbf{Token Embedding.} 
	Before feeding the unmasked point patches into the Transformer-based encoder, we transform them into tokens via a simple PointNet, which consists of an MLP and a max pooling layer. Specifically, we introduce the embedded visible tokens $\T'_{vis} \in \mathbb{R}^{(n-m) \times d}$ as:
	\begin{equation}
	\T'_{vis} = \mathrm{MaxPooling}(\mathrm{MLP}(\P'_{vis})),
	\end{equation}
	where the MLP is applied to the feature dimension while the max pooling is conducted across the $k$ neighboring points.
	
	\textbf{Encoder.}  The encoder is built with the standard Transformer blocks \cite{vaswani2017attention}.  
	Following the common practice, we only feed the visible tokens into the encoder to reduce the computational complexity \cite{pang2022masked}.
	We add the input tokens with the token-wise PE in each Transformer block because the coordinates in point patches are normalized. Finally, the encoded tokens $\T'_{e} \in \mathbb{R}^{(n-m) \times d}$ are introduced as:
	\begin{equation} \label{Equ:transformer_encoder}
	\T'_{e} = \mathrm{E}(\T'_{vis}, \T_{pe}^e(\C'_{vis})),
	\end{equation}
	where $\T_{pe}^e: \mathbb{R}^{(n-m)\times 3} \to \mathbb{R}^{(n-m)\times d}$ is constructed with a learnable MLP and it maps coordinates of patch centers to the feature dimension $d$.
	
	\textbf{Decoder.}
	Motivated by the methods that reconstruct the complete point cloud in a patch-wise manner \cite{yu2021point,pang2022masked}, we promote to decompose the reconstruction of the whole point cloud $\X$ into the reconstructions of detailed local patches and rough global shape.
	Given the encoded visible tokens $\T'_{e}$ and the learnable mask tokens $\T_m \in \mathbb{R}^{m \times d}$, the patch decoder $\mathrm{D}_{patch}$ predicts the decoded mask tokens $\T_d \in \mathbb{R}^{m \times d}$ as:
	\begin{equation} \label{Equ:transformer_decoder}
	\T_{d} = \mathrm{D}_{patch}(\T'_{e}, \T_m, \T_{pe}^d(\C)),
	\end{equation}
	where we duplicate the $d$-dimensional learnable mask token to construct $\T_m$ and the decoder PE, \ie, $\T_{pe}^d: \mathbb{R}^{n\times 3} \to \mathbb{R}^{n\times d}$ share the same structure as the encoder PE of $\T_{pe}^e$. We adopt two separate PEs for encoder and patch decoder since they respectively receive transformed patch centers $\C'_{vis}$ and vanilla patch centers $\C$. The patch decoder $\mathrm{D}_{patch}(\cdot)$ is also constructed with the standard Transformer \cite{vaswani2017attention} but with fewer blocks compared to the encoder. Note that $\T'_{e}$ is the encoded tokens of corrupted point cloud patches, while $\T_{pe}^d(\C)$ and $\T_d$ are the PEs and decoded tokens for the uncorrupted patches, respectively. In other words, we reconstruct the uncorrupted tokens from their corrupted counterparts guided by the uncorrupted patch centers $\C$.

	Given the decoded mask tokens $\T_d$, we predict their corresponding point cloud patches via the folding-based decoder head in Eq. (\ref{Equ:folding}):
	\begin{equation}
	\widehat{\P}_{mask} = \mathrm{D}_{fold}(\T_d),
	\end{equation}
	where $\mathrm{D}_{fold}(\cdot): \mathbb{R}^{m\times d} \to \mathbb{R}^{m\times k \times 3}$ decodes each patch feature in $\T_d$ as a point cloud patch.
	For paired point cloud patches in the reconstructed $\widehat{\P}_{mask}$ and the ground truth $\P_{mask}$, we measure the reconstruction quality via the CD loss in Eq. (\ref{Equ:cd}). 
	The reconstruction loss of local patches are averaged over the $m$ masked patches:
	\begin{equation} \label{Equ:loss_local}
	\mathcal{L}_{local} =  \frac{1}{m} \sum_{i=1}^{m} \mathcal{L}_{cd}(\P_{mask}^i, \widehat{\P}_{mask}^i),
	\end{equation}
	where $\P_{mask}^i \in \mathbb{R}^{k \times 3}$ and $\widehat{\P}_{mask}^i \in \mathbb{R}^{k \times 3}$ are the $i$-th point patch in $\P_{mask}$ and $\widehat{\P}_{mask}$, respectively.
	
	It should be noted that the objective $\mathcal{L}_{local}$ in Eq. (\ref{Equ:loss_local}) only measures the reconstruction performance of normalized local patches, given that the rough global shape is fed into the patch decoder via PE, \eg, $\C$. 
	In other words, the PE leaks the positions of the global shape.
	While this is a common practice in existing methods \cite{yu2021point,pang2022masked}, we argue that the reconstruction of global shape is also important for the reconstruction of a complete point cloud, especially when its global information is corrupted by the affine transformation.
	Therefore, we introduce the following global shape reconstruction loss $\mathcal{L}_{global}$ to complement the local patch reconstruction:
	\begin{equation} \label{Equ:loss_global}
	\mathcal{L}_{global} =  \mathcal{L}_{cd}(\widehat{\C}, \C),
	\end{equation}
	where $\widehat{\C} \in \mathbb{R}^{n \times 3}$ contains the predicted patch centers (\eg, rough global shape) from encoded tokens $\T'_e$ in Eq. (\ref{Equ:transformer_encoder}):
	\begin{equation}
	\widehat{\C}= \mathrm{D}_{fc}(\mathrm{Pooling}(\T'_e)),
	\end{equation}
	where $\mathrm{Pooling}(\cdot): \mathbb{R}^{(n-m)\times d} \to \mathbb{R}^{d}$ merges patch features into a global feature 
	and $\mathrm{D}_{fc}(\cdot): \mathbb{R}^{d} \to \mathbb{R}^{n\times 3}$ is the fully-connected decoder detailed in Eq. (\ref{Equ:fc_decoder}).
	
	\textbf{Final Loss.}
	By decomposing the reconstruction of whole point cloud into the reconstructions of local patches in Eq. (\ref{Equ:loss_local}) and global shape in Eq. (\ref{Equ:loss_global}), we have the following final loss function: 
	\begin{equation} \label{Equ:loss_final_transformer}
	\mathcal{L}_{all} = \mathcal{L}_{local} + \lambda \mathcal{L}_{global},
	\end{equation}
	where $\lambda$ is a hyper-parameter balancing the two loss terms.

	\section{Experiments} \label{Sec:experiments}
	We perform extensive experiments on downstream tasks of point cloud learning with various encoder backbones.  The Point-DAE variants with different types of corruptions are first investigated and then we compare our method against current state of the arts on extensive downstream tasks.
	
	\subsection{Self-supervised Pre-training} \label{Subsec:pre-training}
	
	For most downstream tasks except for the 3D object detection, we conduct self-supervised pre-training on the ShapeNet dataset \cite{shapenet2015}, which contains about $51$K samples shared by $55$ categories, following \cite{wang2021unsupervised,yu2021point,pang2022masked}.
	For the 3D object detection, we follow \cite{liu2022masked} to pre-train our model with the ScanNetv2 dataset \cite{dai2017scannet}, which includes $1,201$, $312$, and $100$ scenes for training, validation, and test, respectively.
	We validate our Point-DAE with point cloud encoders of PointNet \cite{qi2017pointnet}, PointNet++ \cite{qi2017pointnet++}, DGCNN \cite{wang2019dynamic}, standard Transformer \cite{yu2021point,pang2022masked}, \textcolor{black}{Point Transformer \cite{zhao2021point}}, and the hierarchical Transformer \cite{zhang2022point}.
	Especially,  for standard Transformer, we construct the encoder and decoder with $12$ and $4$ Transformer blocks, respectively. Each Transformer block contains $384$ dimension and $6$ heads.  For the hierarchical Transformer encoder, we follow \cite{zhang2022point} to adopt multi-scale masking and a hierarchical patch decoder.
	The AdamW optimizer \cite{loshchilov2017decoupled} and the cosine learning rate schedule \cite{loshchilov2016sgdr} are used to train the network. 
	The total training epochs, initial learning rate, and number of points are respectively set as $300$, $0.001$, and $1024$ unless otherwise specified. 
	
	\begin{table}
		\centering
		\caption{Results of different variants of Point-DAE on the ScanObjectNN dataset with the ObjBG setting. The linear probing protocol is used. 
		} \label{Tab:results_corruptions}
		\begin{tabular}{l|l|c|c|c}
			\hline
			& Corruptions & PointNet & PointNet++ & DGCNN  \\
			\hline
			& \textcolor{gray}{ \large $\blacktriangledown $} No-Corruption  & 75.7$\pm$0.9     & 76.2$\pm$0.9       & 77.1$\pm$0.3\\
			\hline
			& \textcolor{blue}{$\clubsuit$} Drop-Global    & 74.4$\pm$0.2  & 77.3$\pm$0.5        & 76.4$\pm$0.2 \\  
			& \textcolor{blue}{$\clubsuit$}  Scan               & 77.0$\pm$0.1  & 78.1$\pm$1.2       & 78.9$\pm$0.3 \\
			& \textcolor{blue}{$\clubsuit$}  Drop-Patch     & 72.7$\pm$0.4   & 80.6$\pm$1.1       & 77.4$\pm$1.0 \\
			\multirow{-4}{*}{\rotatebox{90}{Masking}}  
			& \textcolor{blue}{$\clubsuit$}  Drop-Local     & 73.6$\pm$0.4   & 78.5$\pm$0.7       & 80.8$\pm$0.8\\  
			\hline
			& \textcolor{PineGreen}{$\blacklozenge$} Add-Global     & 73.6$\pm$0.3    & 77.5$\pm$0.6      & 77.3$\pm$0.9  \\  
			& \textcolor{PineGreen}{$\blacklozenge$} Add-Local       & 74.9$\pm$0.7    & 76.5$\pm$0.5      & 77.2$\pm$0.3  \\  
			\multirow{-3}{*}{\rotatebox{90}{Noise}}  
			& \textcolor{PineGreen}{$\blacklozenge$}  Jitter               & 76.0$\pm$0.8   & 76.1$\pm$0.3      & 78.0$\pm$1.4  \\
			\hline
			& \textcolor{red}{$\spadesuit$} Shear                & 75.6$\pm$0.1    & 77.7$\pm$0.1   & 79.4$\pm$0.6 \\
			& \textcolor{red}{$\spadesuit$} Rotate-Z        & 73.9$\pm$0.4    & 77.9$\pm$0.3      & 79.2$\pm$0.9 \\
			& \textcolor{red}{$\spadesuit$} Translate           &75.1$\pm$0.4    & 78.6$\pm$0.1   & 81.2$\pm$0.7    \\
			& \textcolor{red}{$\spadesuit$} Scale                 & 77.2$\pm$0.9   & 79.9$\pm$0.5   & 80.4$\pm$0.5  \\
			& \textcolor{red}{$\spadesuit$} Reflect          & \textbf{78.3}$\pm$0.7    & 81.8$\pm$0.3   & 82.8$\pm$0.3  \\
			& \textcolor{red}{$\spadesuit$} Rotate           & 76.2$\pm$0.1    & 80.6$\pm$0.8   & 82.8$\pm$0.5 \\
			\multirow{-7}{*}{\rotatebox{90}{Affine Trans.}}  
			& \textcolor{red}{$\spadesuit$}  Affine              & 77.4$\pm$0.7     & 84.9$\pm$1.0   & 84.4$\pm$0.3  \\
			\hline
			&   Affine && &  \\
			&  \textcolor{cyan}{$\bigstar$} + Drop-Patch      & 74.8$\pm$0.1   & 86.0$\pm$0.6               &  84.8$\pm$0.1           \\ 
			\multirow{-3}{*}{\rotatebox{90}{Comb.}}  
			&  \textcolor{cyan}{$\bigstar$} + Drop-Local     & 75.6$\pm$0.5  & \textbf{87.0}$\pm$0.7   & \textbf{85.5}$\pm$0.7 \\
			\hline
			\hline
			& \textcolor{Rhodamine}{$\blacksquare$} Supervised  
			&  78.3$\pm$0.2 & 87.0$\pm$0.1  &  86.8$\pm$0.2 \\
			\hline
		\end{tabular}
	\end{table}
	

	\begin{table}[h!] 
	\centering
	\caption{\textcolor{black}{Results (\%) of Point-DAE on the ScanObjectNN dataset with the ObjBG setting under the linear probing protocol, where the `ST' and `PT' represent the standard Transformer \cite{yu2021point,pang2022masked} and Point Transformer \cite{zhao2021point}, respectively.}  } \label{Tab:results_corruptions_transformer}
	\begin{tabular}{l|cc}
		\hline
		Corruptions & ST & PT  \\
		\hline
		\textcolor{gray}{ \large $\blacktriangledown $} No-Corruption & 76.9$\pm$0.5 & 77.6$\pm$0.3 \\
		\textcolor{blue}{$\clubsuit$}  Drop-Patch  & 82.7$\pm$0.7 & 83.2$\pm$0.3 \\
		\textcolor{red}{$\spadesuit$} Affine  & 79.8$\pm$0.5         & 81.9$\pm$0.6  \\
		\textcolor{cyan}{$\bigstar$}  Affine + Drop-Patch & \textbf{84.6}$\pm$0.6 & \textbf{86.1}$\pm$0.3 \\
		\hline
	\end{tabular}
\end{table}
	
	\subsection{Point-DAE Variants with Different Corruptions} \label{Subsec:results_pointdae_variants}
	
	We first study the Point-DAE variants with seminal non-Transformer encoders, \eg, PointNet \cite{qi2017pointnet}, PointNet++ \cite{qi2017pointnet++}, and DGCNN \cite{wang2019dynamic}, and then verify the findings with Transformer encoders.
	Specifically, we adopt a light-weighted PointNet architecture by removing the transformation modules since the transformation modules (\ie, T-Net) in PointNet are incompatible with the affine transformation corruption (refer to Sec. \ref{Subsec:analyses} for more details).
	Besides the Point-DAE variants induced by the $14$ analyzed corruptions, we introduce the vanilla autoencoder (\ie, Point-DAE with `No-Corruption' input) as the baseline, and explore the Point-DAE variants with combined corruptions. 
	To compare Point-DAE with supervised pre-training, we introduce the `Supervised' setting by learning the encoder with manual category labels.
	We run each variant with three random seeds and report the mean and standard deviation of the results. 
	We adopt the same experimental setting across different Point-DAE variants to ensure a fair comparison. 
	
	Results of Point-DAE variants with different corruption types are illustrated Tab. \ref{Tab:results_corruptions}. Firstly, different corruption types lead to different pretext tasks, resulting in a significant divergence on the downstream performance. For example, masking points nonuniformly (\eg, Drop-Local) typically leads to better performance over uniform point masking (\eg, Drop-Global). Adding noise is a less effective corruption operation, while the affine transformation corruption boosts the results significantly. Combining the revealed affine transformation corruption and the popular masking one leads to consistent improvement with the backbones of PointNet++ and DGCNN, validating their complementarity. Secondly, the same input corruption may lead to opposite effects for different model architectures. For instance, the popular local point masking strategy (\eg, Drop-Patch or Drop-Local) leads to different results with different model architectures.
	For PointNet++ and DGCNN, which are designed with local perception modules, randomly masking the local part of the input introduces explicit regularization, benefiting the extraction and aggregation of local features. Therefore, more effective feature representation can be learned with corruptions of local point masking. On the contrary, masking local points brings a significant negative impact on PointNet, where the global information is overly weighted than local knowledge. Interestingly, the affine transformation corruption brings consistent improvement, revealing that enhancing the global shape perception is important for all model architectures.
	
	The general message in Tab. \ref{Tab:results_corruptions} is the effectiveness of the affine transformation corruption and its complementarity to the popular masking strategy.
	Then, we validate whether these conclusions also hold with the prevalent Transformer backbone, as presented in Tab. \ref{Tab:results_corruptions_transformer}. 
	One may note that affine transformation presents advantages with the non-Transformer encoders while the masking corruption (\eg, Drop-Patch) performs better with the Transformer architecture. This could be attributed to the architecture character, where the patch-wise masking and reconstruction strategy fits the patch-wise processing of the Transformer. Nevertheless, the complementarity between the masking and affine transformation holds with both Transformer and non-Transformer encoders.

	\begin{table}[tb] 
		\centering
		\caption{Linear probing results on ModelNet40 dataset. The `S' and `H' in column `Encoder' indicate the standard Transformer without any special design and the Transformer with hierarchical structure, respectively.}   \label{Tab:modelnet40}
		\begin{subtable}{0.99\linewidth}
			\centering
			\begin{tabular}{l|ccc}
				\hline
				Methods                                          &  \#Points & PointNet & DGCNN   \\
				\hline
				Self-Contrast \cite{du2021self}               & 1K & --      & 89.6  \\ 
				OcCo \cite{wang2021unsupervised}         & 1K   & 88.7  & 89.2   \\  
				Jigsaw \cite{sauder2019self}                   & --   & 87.3  & 90.6   \\  
				Poursaeed \emph{et al.} \cite{poursaeed2020self} & 1K & 88.6  & 90.8 \\
				STRL    \cite{huang2021spatio}               & 2K  & 88.3  & 90.9   \\  
				CrossPoint \cite{afham2022crosspoint} & 1K  & 89.1  & 91.2   \\  
				Point-DAE (Ours)              & 1K & \textbf{89.4}  & \textbf{92.1}\\
				\hline
			\end{tabular}
			\caption{Non-Transformer backbones.} 
		\end{subtable}
		\begin{subtable}{0.99\linewidth}
			\centering
			\begin{tabular}{l|ccc}
				\hline
				Methods & \#Points  & Encoder & Accuracy (\%) \\
				\hline
				Point-MAE \cite{pang2022masked}   & 1K  & S & 90.6 \\
				Point-DAE (Ours)                           &1K  & S & \textbf{92.2} \\
				\hline
				Point-M2AE \cite{zhang2022point}   & 1K  & H & 92.9 \\
				Point-DAE (Ours)                           &1K  & H & \textbf{93.1} \\
				\hline
			\end{tabular}
			\caption{Transformer backbones.} 
		\end{subtable}
	\end{table}
	
	\begin{table}[htb] 
		\centering
		\caption{Classification results on ScanObjectNN dataset. }
		\label{Tab:scanobjectnn}
		\begin{subtable}{0.99\linewidth}
			\centering
			\begin{tabular}{l|ccc}
				\hline
				Supervised References  & \multicolumn{2}{c}{OBJ-BG Results}   \\
				\hline
				PointNet++ \cite{qi2017pointnet++}   & \multicolumn{2}{c}{82.3} \\
				PointCNN \cite{li2018pointcnn}  & \multicolumn{2}{c}{86.1} \\
				\hline
				\hline
				SSL Methods  & PointNet & DGCNN  \\
				\hline
				\multicolumn{3}{c}{\textbf{Fine-tuning Protocol}}\\
				\hline
				From Scratch                                          &73.3 & 82.4      \\
				+ Jigsaw \cite{sauder2019self}                 & 76.5 & 82.7   \\
				+ OcCo \cite{wang2021unsupervised}      & 80.0 & 83.9   \\
				+ PointVST \cite{zhang2022self}              & -- & 89.3 \\
				+ IAE \cite{yan2022implicit}                    & --    & 90.2   \\
				+ Point-DAE (Ours)                                 & \textbf{80.2} &  \textbf{92.1}  \\
				\hline
				
				\hline
				\multicolumn{3}{c}{\textbf{Linear Probing Protocol}}\\
				\hline
				Jigsaw \cite{sauder2019self}                  & 55.2  & 59.5   \\
				OcCo \cite{wang2021unsupervised}      & 69.5  & 78.3   \\
				CrossPoint \cite{afham2022crosspoint}  & 75.6 & 81.7   \\
				Point-DAE (Ours)                                  &  \textbf{78.1} & \textbf{87.9}  \\
				\hline
			\end{tabular}
			\caption{Non-Transformer backbones}
		\end{subtable}
		\begin{subtable}{0.99\linewidth}
			\centering
			\begin{tabular}{lccc}
				\hline
				Methods & OBJ-BG & OBJ-ONLY & PB-T50-RS  \\
				\hline
				\multicolumn{4}{c}{\textbf{Fine-tuning Protocol}}\\
				\hline
				Standard Transformer \cite{yu2021point}  & 79.86 & 80.55 & 77.24 \\
				+ OcCo \cite{wang2021unsupervised,yu2021point}  & 84.85 & 85.54 & 78.79 \\
				+ Point-BERT \cite{yu2021point} & 87.43 & 88.12 & 83.07 \\
				+ Point-GAME \cite{liu2023pointgame} & 88.99 & 87.95 & 83.79 \\
				+ MaskPoint \cite{liu2022masked}     & 89.3 & 89.7 & 84.6  \\
				+ Point-MAE \cite{pang2022masked} & 90.02 & 88.29 & 85.18  \\
				+ Point-CAM \cite{szachniewicz2023self}     & 90.36 & 88.35 & 84.49 \\
				+ TAP \cite{wang2023tap}              & 90.36 & 89.50 & 85.67 \\
				+ Joint-MAE \cite{guo2023joint}       & 90.94   &  88.86 & 86.07  \\ 
				+ MAE3D \cite{jiang2022masked}         & -- & -- & 86.20 \\
				+ Point2vec \cite{zeid2023point2vec}    & 91.2 & 90.4 & 87.5 \\
				+ ACT \cite{dong2022autoencoders}             & 93.29   & 91.91 & 88.21 \\
				+ Point-DAE (Ours)                            & \textbf{93.63}   &  \textbf{92.77}   & \textbf{88.69}  \\
				\hline
				Hierarchical Transformer \cite{zhang2022point}  & 82.31  & 81.57  & 79.35 \\
				+ Point-M2AE \cite{zhang2022point}  &  91.22  & 88.81 & 86.43 \\
				+ IAE \cite{yan2022implicit}               & 92.5   & 91.6  & 88.2 \\
				+ Point-DAE (Ours)                          &  \textbf{93.86}  & \textbf{93.07} & \textbf{89.31}  \\
				\hline
				\multicolumn{4}{c}{\textbf{Linear Probing Protocol}}\\
				\hline
				Point-MAE \cite{pang2022masked} & 80.13  & 80.62  & 70.41 \\
				Point-DAE (Ours) & \textbf{84.62} & \textbf{84.73} & \textbf{74.62} \\
				\hline
			\end{tabular}
			\caption{Transformer backbones}
		\end{subtable}
	\end{table}
	
	\subsection{Results on Downstream Tasks} \label{Subsec:fine-tune}
	
	We adopt the following two protocols to validate the pre-trained encoder on downstream tasks:
	\begin{itemize}
		\item Fine-tuning protocol, where we fine-tune both the pre-trained encoder and the randomly initialized classification head on downstream tasks.
		\item Linear probing protocol, where we freeze the pre-trained encoder and learn a linear support vector machine (SVM) classifier on the extracted features.
	\end{itemize}
	In the following studies, results of competitors are either copied from the original papers or quoted from \cite{wang2021unsupervised,afham2022crosspoint}.

	\textbf{Object Classification.}
	We validate our method on object classification datasets of ModelNet40 and ScanObjectNN.
	The widely-used ModelNet40 dataset \cite{wu20153d} includes about $12K$ synthetic objects shared by $40$ categories, while the challenging ScanObjectNN dataset \cite{uy2019revisiting} consists of about $15K$ scanned point cloud instances shared by $15$ classes.
	As shown in Tab. \ref{Tab:modelnet40}, Point-DAE consistently outperforms the competitors on the ModelNet40 dataset with various encoder backbones.
	On the challenging ScanObjectNN dataset, our Point-DAE shows more considerable advantages, as illustrated in Tab. \ref{Tab:scanobjectnn}. 
	
	Our method also achieves better performance with more powerful backbones, as shown by the comparison between PointNet and DGCNN, and standard and hierarchical Transformers.
	The advantages with different backbones, on different datasets, and under different evaluation protocols fully justify the effectiveness and generality of our method.
	We provide the ModelNet40 results \cite{wu20153d} under the fine-tuning protocol in the \textbf{Supplementary Material} since performance on this simple task is already saturated (\ie, around 94\% accuracy).

	\begin{table*} 
		\centering
		\caption{Robustness to common corruptions on the ModelNet-C test dataset. } \label{Tab:robustness}
		\begin{tabular}{l|c|c|ccccccc}
			\hline
			Methods      & OA $\uparrow$      & mCE $\downarrow$   & Scale   & Jitter   & Drop-G   & Drop-L   & Add-G   & Add-L   & Rotate \\
			\hline
			DGCNN (From Scratch)              & 0.926 & 1.000  & 1.000  & 1.000 & 1.000     & 1.000      & 1.000    & 1.000   & 1.000  \\
			+ OcCo \cite{wang2021unsupervised} & 0.922 & 1.047  & 1.606  & \textbf{0.652} & \textbf{0.903}      & 1.039     &  1.444   & 0.847   & 0.837 \\
			+ Point-DAE (Ours) & \textbf{0.933} & \textbf{0.835} & \textbf{0.957} & 0.883 & 0.944    & \textbf{0.841}     &   \textbf{0.668}  & \textbf{0.815}   & \textbf{0.736}  \\
			\hline
		\end{tabular}
	\end{table*}
	
	\vspace{0.1cm}
	\textbf{Robustness Testing.} 
	We study the model robustness to sample corruptions on the ModelNet-C test set \cite{ren2022benchmarking}, which is constructed by applying seven atomic corruptions to the vanilla ModelNet40 test set. 
	Among the seven studied corruptions, corruptions similar to Scale, Rotate, and Drop-L have been utilized in the pre-training stage of Point-DAE, while the other four corruptions of Jitter, Drop-G, Add-G, and Add-L have not been exposed. 
	In the fine-tuning stage, we strictly follow the evaluation protocol of \cite{ren2022benchmarking} and only use conventional augmentations of scaling and translation.
	Besides the overall accuracy (OA) on the clean test set, we report the mean Corruption Error (mCE) for the robustness evaluation.
	As illustrated in Tab. \ref{Tab:robustness}, Point-DAE outperforms the vanilla DGCNN on the clean test set and achieves improved robustness to all the seven investigated corruptions.
	Especially the results on Jitter, Drop-G, Add-G, and Add-L corruptions validate its improved generalization capability to unseen corruptions.
	
	\textbf{Robustness to SO(3) Rotation.}
	Although the model robustness to rotations has been preliminarily investigated with the ModelNet-C test set, it is restricted to small angle ranges \cite{ren2022benchmarking}. 
	In this analysis, we investigate the Point-DAE in the more challenging setting of arbitrary SO(3) rotations \cite{poulenard2019effective,rao2019spherical,kim2020rotation}.
	Specifically, we adopt two experiment settings with the DGCNN backbone following the common practice \cite{poulenard2019effective,rao2019spherical,kim2020rotation}. 
	We randomly rotate the training and test data around the $z$-axis in the `$z/z$' setting.
	In the `SO(3)/SO(3)' setting, training and test data are randomly rotated around the $xyz$-axes.
	As illustrated in Tab. \ref{Tab:so3_testing}, Point-DAE improves over the DGCNN baseline on both settings, validating the improved rotation robustness to arbitrary SO(3) rotations.
	
	\begin{table}
		\centering
		\caption{Classification results with different rotation settings. All comparing methods adopt `xyz' input only. } \label{Tab:so3_testing}
		\begin{tabular}{l|cc}
			\hline
			Methods &  $z/z$  & SO(3)/SO(3) \\
			\hline
			\multicolumn{3}{c}{ModelNet40}   \\
			\hline
			\multicolumn{3}{c}{Rotation-specific Methods} \\
			SPHNet \cite{poulenard2019effective} & 87.7  & 87.2  \\
			SFCNN \cite{rao2019spherical} &  91.4                & \textbf{90.1}                 \\
			RI-GCN \cite{kim2020rotation} & 89.5                 &  89.5                \\
			\hline
			DGCNN  (From Scratch)                   &  90.1    & 88.3  \\
			+ Point-DAE (Ours)  &  \textbf{91.6}    & 89.3  \\
			\hline
			\hline
			\multicolumn{3}{c}{ObjBG setting of ScanObjectNN} \\
			\hline 
			DGCNN (From Scratch)                    & 79.9   & 79.4  \\
			+ Point-DAE (Ours) &  \textbf{88.3} &   \textbf{85.6}  \\
			\hline
		\end{tabular}
	\end{table}
	
	\vspace{0.1cm}
	\textbf{Part Segmentation}
	We also validate Point-DAE on the part segmentation task with the ShapeNetPart dataset, which includes $16,881$ samples shared by $16$ classes. As illustrated in Tab. \ref{Tab:partseg}, our method outperforms previous methods by $+0.7\%$, $+0.5\%$, $+0.3\%$ with backbones of PointNet, DGCNN, and standard Transformer, respectively.

	\begin{table} [tb] 
		\centering
		\caption{Part segmentation results (mean IoU) on the ShapeNetPart dataset.} \label{Tab:partseg}
		\begin{tabular}{lccc}
			\hline
			Methods  & PointNet & DGCNN & Transformer \\ 
			\hline
			From Scratch  & 82.2 & 85.2 & 85.1 \\ 
			+ OcCo \cite{wang2021unsupervised}           & 83.4 & 85.0 & 85.1 \\ 
			+ Jigsaw \cite{sauder2019self}                        &  -- & 85.1 & -- \\ 
			+ PointContrast \cite{xie2020pointcontrast}   &  -- & 85.3 & -- \\ 
			+ CrossPoint \cite{afham2022crosspoint}      & -- & 85.5 & -- \\ 
			+ Chen \emph{et al.} \cite{chen2021shape}  & 84.1 & -- & --\\
			+ Point-BERT \cite{yu2021point}                  & -- & --  & 85.6   \\
			+ MaskPoint \cite{liu2022masked}               & -- & --  & 86.0 \\
			+ Point-MAE \cite{pang2022masked}           & -- & -- & 86.1  \\
			+ Point-DAE (Ours)  & \textbf{84.8} &  \textbf{86.0} & \textbf{86.4}  \\
			\hline
		\end{tabular}
	\end{table}
	
	\vspace{0.1cm}
	\textbf{3D Object Detection}
	We also validate Point-DAE on the 3D objection detection task with the ScanNetv2 dataset \cite{dai2017scannet}, which includes $1,201$, $312$, and $100$ scenes for training, validation, and test, respectively. 	We closely follow \cite{liu2022masked} to conduct experiments for 3D object detection. 
	Specifically, we pre-train our model on the ScanNet dataset, which includes about 2.5 million RGBD scans. Only the geometry information is utilized, while the RGB knowledge is discarded.
	We adopt the `ScanNet-Medium' dataset, which is generated by sampling every $100$ frame from ScanNet and contains around $25$K samples.
	For each instance, we sample $20$K points and cluster them into $2,048$ groups, each containing $64$ points. We randomly mask $75\%$ of the groups as input to the Transformer encoder. 
	The 3DETR model is adopted for the 3D objection detection task, where the encoder includes 3-layer self-attention blocks, and the decoder is composed of 8-layer cross-attention blocks. 
	When fine-tuning, we strictly follow the 3DETR strategy, except that we initialize the encoder with the pre-trained weights. 
	As illustrated in Tab. \ref{Tab:detection}, our method significantly improves over the 3DETR baseline (+2.0AP$_{25}$ and +4.9AP$_{50}$) and outperforms its closest competitor, Point-MAE, validating the effectiveness of Point-DAE on scene understanding.
	
	\begin{table} [tb] 
		\centering
		\caption{3D objection results on ScanNet validation set.} \label{Tab:detection}
		\begin{tabular}{lcc}
			\hline
			Methods  & AP$_{25}$ & AP$_{50}$  \\ 
			\hline
			VoteNet \cite{qi2019deep}        & 58.6 & 33.5 \\
			+ PointContrast \cite{xie2020pointcontrast} &  59.2 & 38.0 \\
			+ DepthContrast \cite{zhang2021self} & 61.3 & --  \\
			\hline
			3DETR \cite{misra2021end}          & 62.1 & 37.9  \\
			+ MaskPoint \cite{liu2022masked} & 63.4 & 40.6  \\
			+ Point-MAE \cite{pang2022masked} & 63.3 & 40.4   \\  
			+ Point-DAE (Ours)  &  \textbf{64.1} & \textbf{42.8}   \\
			\hline
		\end{tabular}
	\end{table}

	\begin{table}[htb] 
		\centering
		\caption{Few-shot classification results with $1$K input points.
		} \label{Tab:few_shot_all}
		\begin{subtable}{0.99\linewidth}
			\centering
			\begin{tabular}{lcccc}
				\hline
				& \multicolumn{2}{c}{\textbf{5-way}} & \multicolumn{2}{c}{\textbf{10-way}} \\
				\cmidrule(r){2-3} 	\cmidrule(r){4-5}
				& 10-shot & 20-shot & 10-shot & 20-shot \\
				\hline
				PointNet \cite{qi2017pointnet}  & 52.0$\pm$3.8 & 57.8$\pm$4.9 & 46.6$\pm$4.3 & 35.2$\pm$4.8 \\
				+ Jigsaw \cite{sauder2019self}  & 66.5$\pm$2.5 & 69.2$\pm$2.4 & 56.9$\pm$2.5 & 66.5$\pm$1.4 \\
				+ cTree \cite{sharma2020self}  & 63.2$\pm$3.4 & 68.9$\pm$3.0 & 49.2$\pm$1.9 & 50.1$\pm$1.6 \\
				+ OcCo \cite{wang2021unsupervised}   & 89.7$\pm$1.9 & 92.4$\pm$1.6 & 83.9$\pm$1.8 & 89.7$\pm$1.5 \\
				+ CrossPoint \cite{afham2022crosspoint} & 90.9$\pm$4.8 & 93.5$\pm$4.4 & 84.6$\pm$4.7 & 90.2$\pm$2.2 \\
				+ Point-DAE (Ours) & \textbf{93.0}$\pm$3.7 & \textbf{94.9}$\pm$3.3 & \textbf{86.7}$\pm$5.8 & \textbf{92.1}$\pm$4.6 \\
				\hline
				DGCNN \cite{wang2019dynamic}    & 31.6$\pm$2.8 & 40.8$\pm$4.6 & 19.9$\pm$2.1  &  16.9$\pm$1.5  \\
				+ Jigsaw \cite{sauder2019self} & 34.3$\pm$1.3 & 42.2$\pm$3.5 & 26.0$\pm$2.4 & 29.9$\pm$2.6 \\
				+ cTree \cite{sharma2020self}  & 60.0$\pm$2.8 & 65.7$\pm$2.6 & 48.5$\pm$1.8 & 53.0$\pm$1.3    \\
				+ OcCo \cite{wang2021unsupervised}  & 90.6$\pm$2.8 & 92.5$\pm$1.9 & 82.9$\pm$1.3 & 86.5$\pm$2.2 \\
				+ CrossPoint \cite{afham2022crosspoint} & 92.5$\pm$3.0 & 94.9$\pm$2.1 & 83.6$\pm$5.3 & 87.9$\pm$4.2  \\
				+ Point-DAE (Ours)      & \textbf{96.7}$\pm$2.5 & \textbf{97.7}$\pm$1.6 & \textbf{93.0}$\pm$4.8 & \textbf{95.6}$\pm$2.6 \\
				\hline
				Transformer \cite{yu2021point} & 87.8$\pm$5.2 & 93.3$\pm$4.3 & 84.6$\pm$5.5 & 89.4$\pm$6.3 \\
				+ OcCo \cite{yu2021point} & 94.0$\pm$3.6 & 95.9$\pm$2.3 & 89.4$\pm$5.1 & 92.4$\pm$4.6 \\
				+ Point-BERT \cite{yu2021point} & 94.6$\pm$3.1 & 96.3$\pm$2.7 & 91.0$\pm$5.4 & 92.7$\pm$5.1 \\
				+ MaskPoint \cite{liu2022masked}     & 95.0$\pm$3.7 & 97.2$\pm$1.7 & 91.4$\pm$4.0 & 93.4$\pm$3.5 \\
				+ Point-MAE \cite{pang2022masked} & 96.3$\pm$2.5 & 97.8$\pm$1.8 & 92.6$\pm$4.1 & 95.0$\pm$3.0 \\
				+ Point-DAE (Ours)                          &  \textbf{96.8}$\pm$2.4 & \textbf{98.3}$\pm$1.5 & \textbf{93.2}$\pm$4.6 & \textbf{95.6}$\pm$3.2 \\
				\hline
			\end{tabular}
			\caption{ModelNet40} 
		\end{subtable}
		\begin{subtable}{0.99\linewidth}
			\centering
			\begin{tabular}{lcccc}
				\hline
				& \multicolumn{2}{c}{\textbf{5-way}} & \multicolumn{2}{c}{\textbf{10-way}} \\
				\cmidrule(r){2-3} 	\cmidrule(r){4-5}
				& 10-shot & 20-shot & 10-shot & 20-shot \\
				\hline
				PointNet \cite{qi2017pointnet}     & 57.6$\pm$2.5  & 61.4$\pm$2.4 & 41.3$\pm$1.3 & 43.8$\pm$1.9 \\
				+ Jigsaw \cite{sauder2019self}               & 58.6$\pm$1.9  &  67.6$\pm$2.1 & 53.6$\pm$1.7 & 48.1$\pm$1.9 \\
				+ cTree \cite{sharma2020self}                & 59.6$\pm$2.3 & 61.4$\pm$1.4 & 53.0$\pm$1.9 & 50.9$\pm$2.1 \\
				+ OcCo \cite{wang2021unsupervised}      & 70.4$\pm$3.3 & 72.2$\pm$3.0 & 54.8$\pm$1.3 & 61.8$\pm$1.2 \\
				+ CrossPoint \cite{afham2022crosspoint} & 68.2$\pm$1.8 & 73.3$\pm$2.9 & 58.7$\pm$1.8 & 64.6$\pm$1.2 \\
				+ Point-DAE (Ours) & \textbf{71.3}$\pm$6.9 & \textbf{74.5}$\pm$5.7      & \textbf{59.6}$\pm$3.7  & \textbf{66.2}$\pm$4.5  \\
				\hline
				DGCNN \cite{wang2019dynamic}    & 62.0$\pm$5.6 & 67.8$\pm$5.1 & 37.8$\pm$4.3 & 41.8$\pm$2.4 \\
				+ Jigsaw  \cite{sauder2019self}                 & 65.2$\pm$3.8 & 72.2$\pm$2.7 & 45.6$\pm$3.1 & 48.2$\pm$2.8 \\
				+ cTree  \cite{sharma2020self}                 & 68.4$\pm$3.4 & 71.6$\pm$2.9 & 42.4$\pm$2.7 & 43.0$\pm$3.0  \\
				+ OcCo   \cite{wang2021unsupervised}    & 72.4$\pm$1.4 & 77.2$\pm$1.4 & 57.0$\pm$1.3 & 61.6$\pm$1.2  \\
				+ CrossPoint \cite{afham2022crosspoint} & 74.8$\pm$1.5 & 79.0$\pm$1.2 & 62.9$\pm$1.7 & 73.9$\pm$2.2 \\
				+ Point-DAE (Ours)            & \textbf{84.5}$\pm$4.5  & \textbf{88.2}$\pm$5.3 & \textbf{76.4}$\pm$1.9  & \textbf{81.6}$\pm$2.4 \\
				\hline
				Transformer       & 62.6$\pm$9.6 & 69.8$\pm$8.9 & 48.6$\pm$3.0 & 57.8$\pm$4.0    \\
				+ Point-MAE \cite{pang2022masked}         & 74.5$\pm$5.7  & 83.9$\pm$7.7 & 69.0$\pm$3.7 & 77.6$\pm$3.8  \\
				+ Point-DAE (Ours) & \textbf{84.8}$\pm$5.7 & \textbf{88.1}$\pm$7.2 & \textbf{76.4}$\pm$3.8 & \textbf{81.0}$\pm$4.7 \\
				\hline
			\end{tabular}
			\caption{OBJ-BG Setting of ScanObjectNN} \label{Tab:few_shot_scanobjectnn}
		\end{subtable}
	\end{table}
	
	\vspace{0.1cm}
	\textbf{Few-shot Classification.}
	Learning to perform new tasks with few training samples is vital for pre-training. To validate this ability, we conduct few-shot classification on datasets of ScanObjectNN and ModelNet40 under the $i$-way, $j$-shot setting, where $i$ is the number of randomly selected classes and $j$ is the number of training samples in each class. We set $i \in \{5,10\}$ and $j \in \{10, 20\}$ and conduct $10$ independent experiments in each setting for the performance report. 
	As presented in Tab. \ref{Tab:few_shot_all}, our method consistently outperforms its competitors with different backbones, 
	Our Point-DAE consistently outperforms its competitors, while the advantages on more challenging tasks (\eg, the ScanObjectNN dataset) are more significant. Especially, our Point-DAE considerably outperforms the CrossPoint by $13.5\%$ in the hardest $10$-way, $10$-shot setting with a DGCNN backbone on the ScanObjectNN dataset.

	\subsection{Analyses and Discussions} \label{Subsec:analyses}
	The following analyses are conducted on the OBJ-BG setting of ScanObjectNN dataset under the linear probing protocol unless otherwise specified.
	
\begin{figure}[tb]
		\begin{subfigure}[t]{.49\linewidth}
			\centering\includegraphics[width=.9\linewidth]{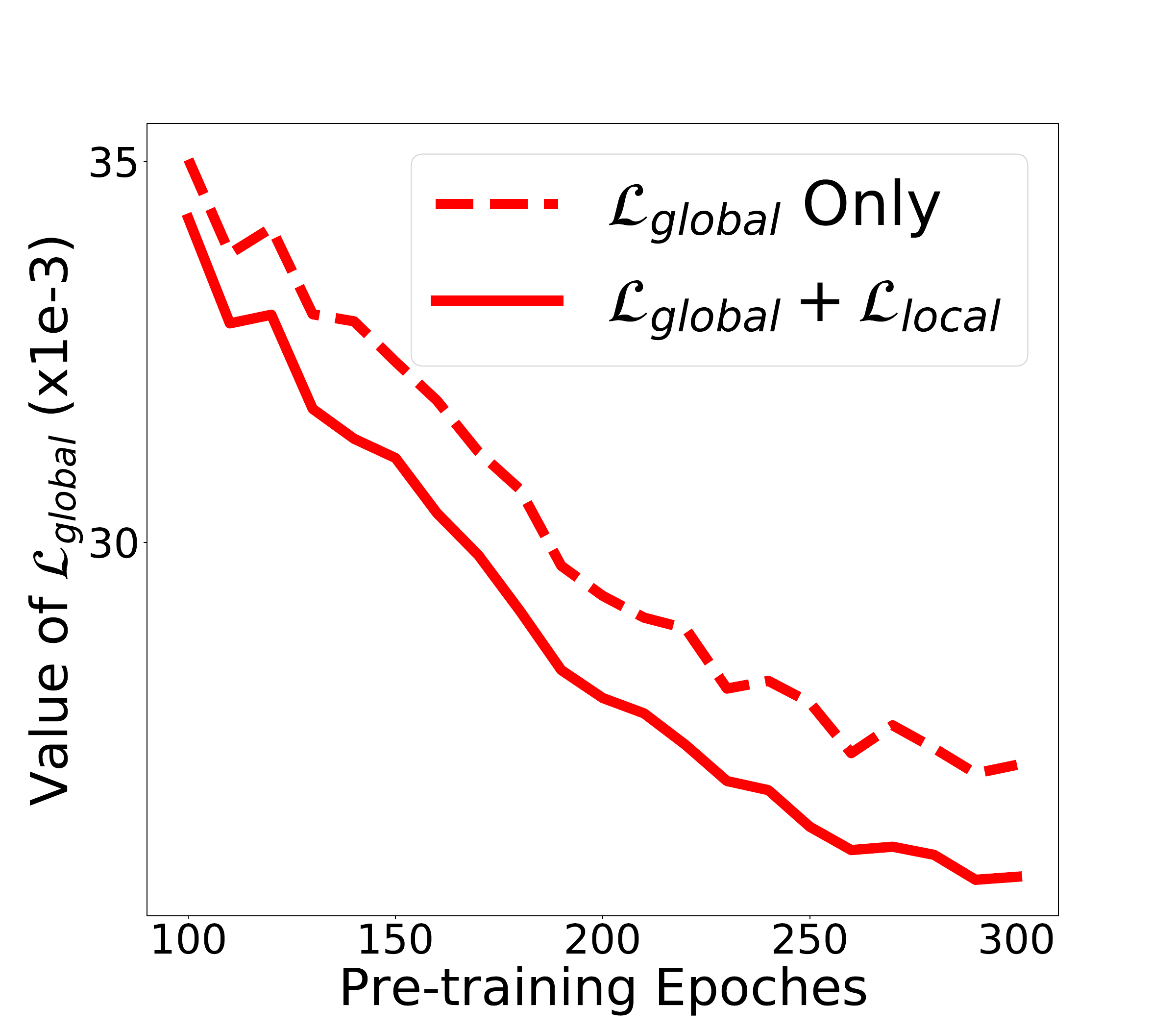}
			\caption{$\mathcal{L}_{global}$}
		\end{subfigure}
		\begin{subfigure}[t]{.49\linewidth}
			\centering\includegraphics[width=.9\linewidth]{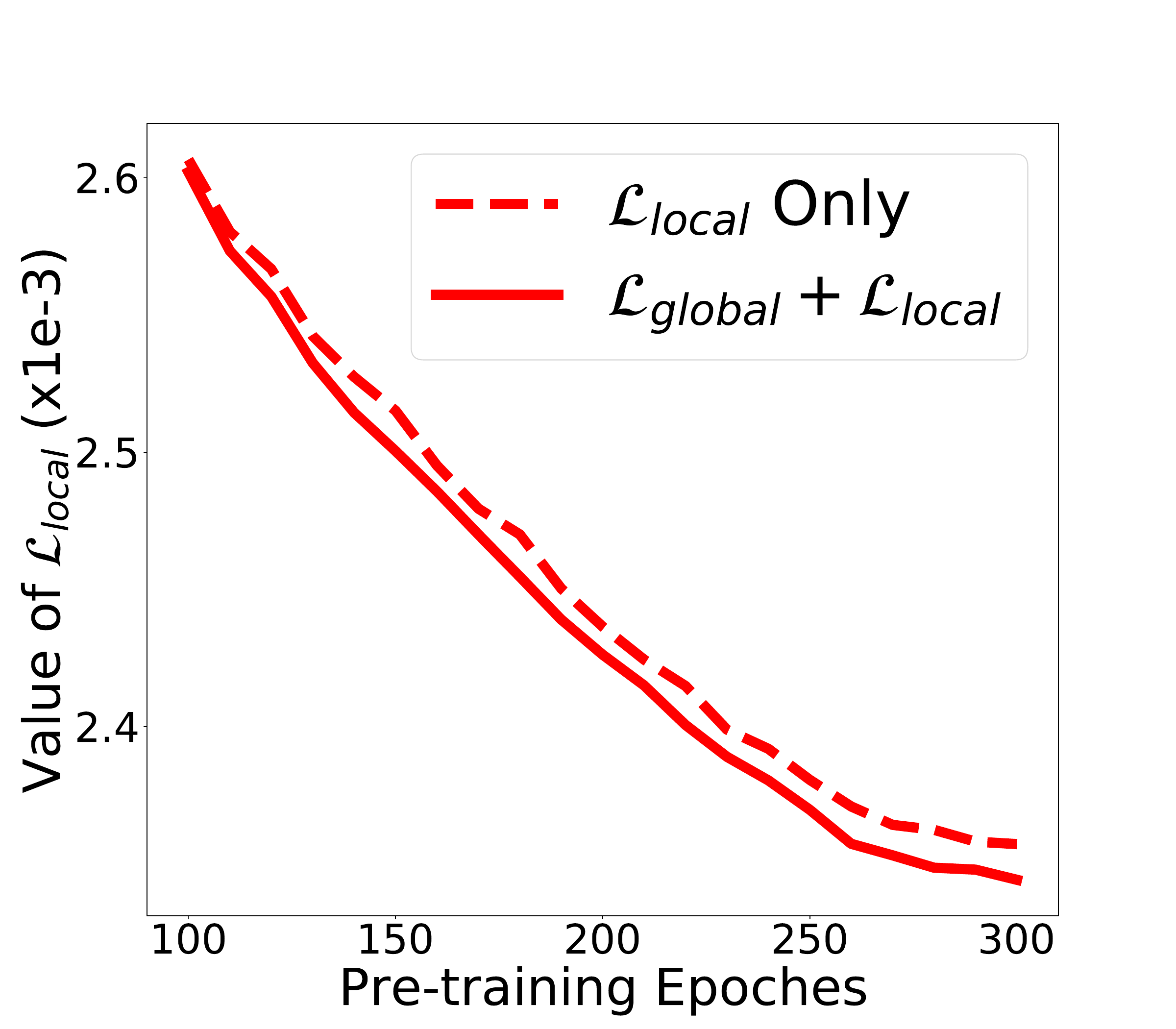}
			\caption{$\mathcal{L}_{local}$}
		\end{subfigure}
		\caption{Loss curves of (a) $\mathcal{L}_{global}$ and (b) $\mathcal{L}_{local}$ with different learning objectives.
		}\label{Fig:local_global_loss}
	\end{figure}

	\textbf{Reconstruction Decomposition with Transformer.}
	Instead of decomposing the reconstruction of the whole point cloud into the reconstructions of local patches $\mathcal{L}_{local}$ and global shape $\mathcal{L}_{global}$, one may opt to reconstruct the whole point cloud directly by revising Eq. (\ref{Equ:loss_global}) as $\mathcal{L}_{whole} =  \mathcal{L}_{cd}(\widehat{\X}, \X)$, where $\widehat{\X} \in \mathbb{R}^{w\times 3}$ is obtained via the fully-connected decoder $\mathrm{D}_{fc}(\mathrm{Pooling}(\T'_e))$.
	As illustrated in Tab. \ref{Tab:global_local_losses}, decomposing the whole reconstruction loss as two sub-terms introduces considerable improvement (see the comparison between A1 and A4). Among the two sub-terms, the local patch reconstruction loss $\mathcal{L}_{local}$ contributes more under the fine-tuning protocol, while better linear probing performance is achieved with the global shape reconstruction loss $\mathcal{L}_{global}$, suggesting their complementarity. This can also be observed from the loss curves plotted in Fig. \ref{Fig:local_global_loss}. We see that introducing the global shape reconstruction loss $\mathcal{L}_{global}$ facilitates the convergence of the local patch reconstruction $\mathcal{L}_{local}$, and vice versa.
	
	\begin{table}[tb] 
		\centering
		\caption{Classification results (\%) with different objectives.} \label{Tab:global_local_losses}
		\begin{tabular}{c|c|cc|cc}
			\hline
			& $\mathcal{L}_{whole}$ & $\mathcal{L}_{global}$   & $\mathcal{L}_{local}$      &  Linear Probing & Fine-tuning \\
			\hline
			A1&$\large{\checkmark}$ & & & 83.56 & 91.01 \\
			\hline
			A2& & $\large{\checkmark}$ &        &   82.51             &   89.75 \\  
			
			A3& &                                  &  $\large{\checkmark}$        &   80.35              &  91.16 \\
			
			A4&  & $\large{\checkmark}$ &   $\large{\checkmark}$     &   \textbf{84.62}             &   \textbf{93.63}  \\
			\hline     
		\end{tabular}
	\end{table}

\begin{figure}[tb]
		\centering
		\begin{subfigure}[t]{.79\linewidth}
			\centering\includegraphics[width=.96\linewidth]{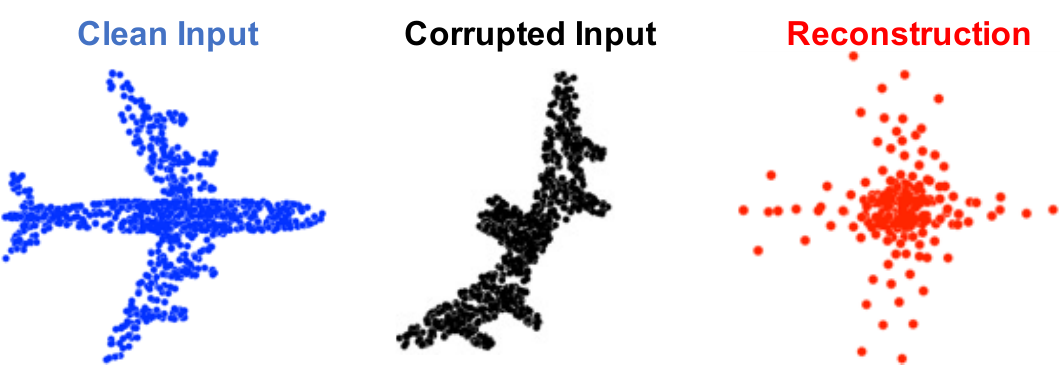}
			\caption{DGCNN}
		\end{subfigure}
		\centering
		\begin{subfigure}[t]{.99\linewidth}
			\centering\includegraphics[width=.96\linewidth]{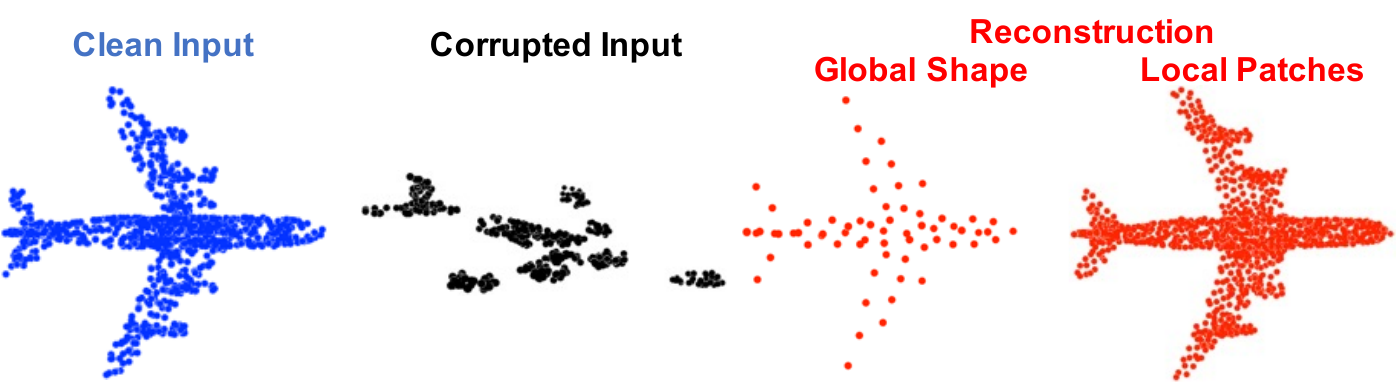}
			\caption{Transformer}
		\end{subfigure}
		\vspace{-0.2cm}
		\caption{Visualization of reconstruction on ShapeNet validation set with backbones of DGCNN and Transformer. More visualization results and discussions can be found in the \textbf{Supplementary Material}.
		}\label{Fig:reconstruct_vis}
	\end{figure}

	\vspace{0.1cm}
	\textbf{Decoder Architectures.}
	While different encoder backbones have been investigated in existing works \cite{wang2021unsupervised,pang2022masked}, there are fewer studies on decoder architectures for the reconstruction-based SSL framework. 
	Here we study the fully-connected decoder \cite{achlioptas2018learning,yuan2018pcn} and the folding-based decoder \cite{yang2018foldingnet}, which are described in Sec. \ref{Subsec:non_transformer}, with Point-DAE framework. 
	As illustrated in Fig. \ref{Fig:reconstruct_vis}, although the reconstruction of a complete point cloud can be achieved with the DGCNN backbone, it only reconstructs the rough global shape due to the challenging corruptions.
	As can be seen in Tab. \ref{Tab:decoders}, the fully-connected decoder shows advantages on global shape reconstruction, which can be validated by comparing B1\&B2, B3\&B5, and B4\&B6. On the contrary, the folding-based decoder is good at local patch reconstruction, as verified by the comparisons between B3\&B4, and  B5\&B6. 
	These findings echo the observations that the fully-connected decoder is good at capturing the global shape geometry, while the folding-based decoder is good at characterizing a smooth local surface \cite{yuan2018pcn}. 
	
	\begin{table}[tb] 
		\centering
		\caption{Results (\%) with different decoder architectures.} \label{Tab:decoders}
		\begin{tabular}{c|cc|c}
			\hline
			& \multicolumn{3}{c}{DGCNN Backbone}         \\
			\hline
			& \multicolumn{2}{c|}{Complete Point Cloud Decoder}    & Acc. (\%) \\
			\hline
			B1 & \multicolumn{2}{c|}{Folding-based}           &  80.0         \\
			B2 & \multicolumn{2}{c|}{Fully-connected}         &   \textbf{85.5} \\
			\hline
			\hline
			&\multicolumn{3}{c}{Transformer Backbone}   \\
			\hline
			&Global Shape Decoder & Local Patch Decoder & Acc(\%)   \\
			\hline
			B3 & Folding-based        & Fully-connected            &       79.0     \\ 
			B4 & Folding-based        & Folding-based              &        81.2    \\ 
			B5 & Fully-connected      & Fully-connected           &         83.8   \\
			B6 & Fully-connected      & Folding-based             &        \textbf{84.6}    \\  
			\hline  
		\end{tabular}
	\end{table}
	
	\begin{figure}[tb]
		\centering\includegraphics[width=.75\linewidth]{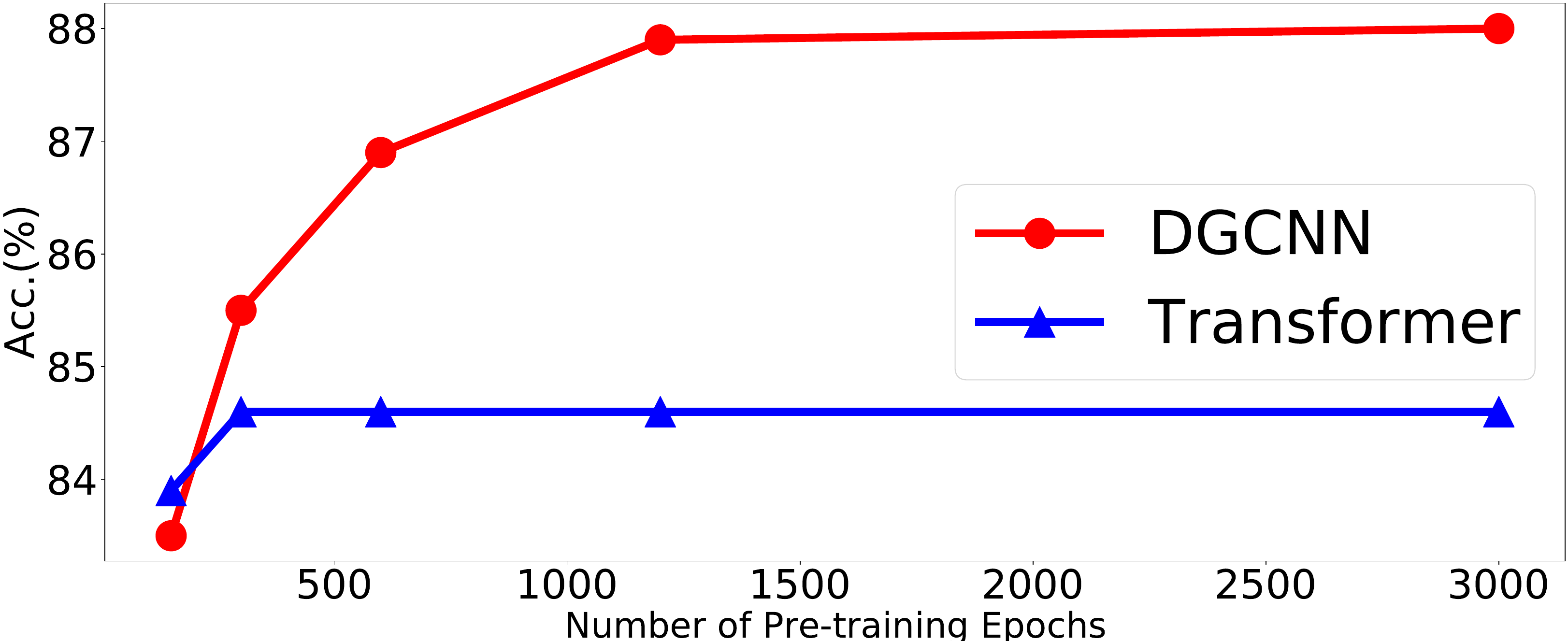}
		\caption{Results (\%) with different number of pre-training epochs.
		}\label{Fig:pre_training_epoch}
	\end{figure}
	
	\vspace{0.1cm}
	\textbf{Masking Variants.}
	Three masking strategies are compared in Tab. \ref{Tab:local_mask_dgcnn} with only masking corruption. In the `View' setting, points occluded in a random camera view are masked \cite{wang2021unsupervised}. 
	We mask fixed-sized KNN clusters and random-sized KNN clusters in the `Fixed-sized' and `Random-sized' settings, as detailed in Sec. \ref{Subsec:transformer_methods} and Sec. \ref{Subsec:non_transformer}, respectively. 
	The `Random-sized' setting presents a clear advantage with the DGCNN backbone. 
	However, this setting is incompatible with the Transformer backbones, where input patches are `Fixed-sized' point clusters.
	
	\begin{table} \small
		\centering
		\caption{Results (\%) with different local masking methods.} \label{Tab:local_mask_dgcnn}
		\begin{tabular}{l|ccc}
			\hline
			Masking & View \cite{wang2021unsupervised} & Fixed-sized \cite{pang2022masked} & Random-sized \\
			\hline
			Acc (\%) & 78.9 & 79.5       & \textbf{81.3}  \\
			\hline
		\end{tabular}
	\end{table}

	%
	\begin{figure*}
		\centering
		\includegraphics[width=0.96\linewidth]{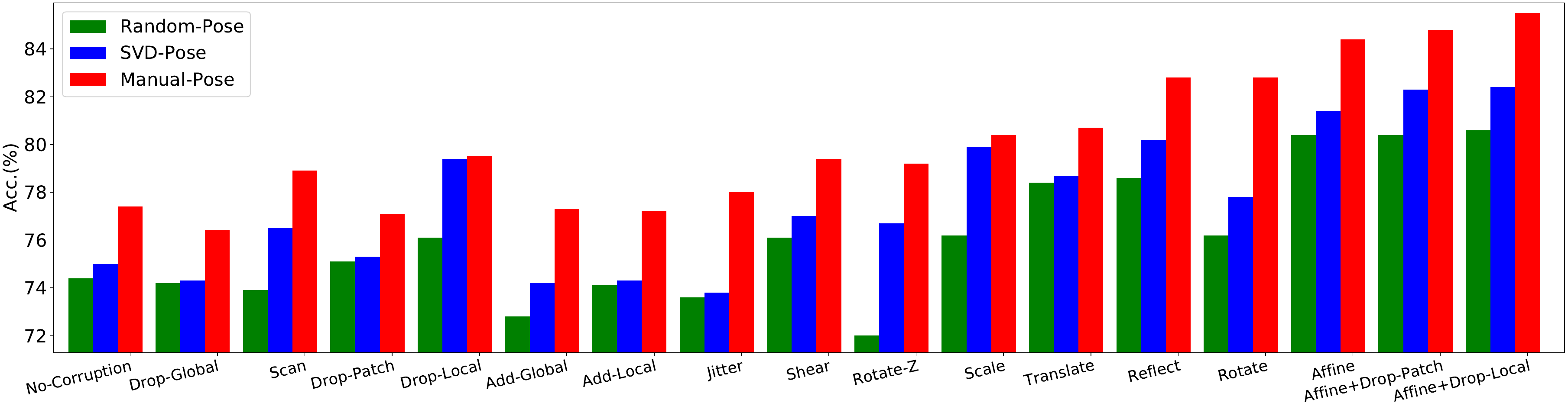}
		\caption{Results of Point-DAE with different pre-training datasets with and without manual poses. Experiments are conducted on the ObjBG setting of ScanObjectNN dataset under the linear classification protocol with the DGCNN backbone.
		}\label{Fig:dataset_variants}
	\end{figure*}

%
	
	\vspace{0.1cm}
	\textbf{Pre-training Epochs.}  
	As shown in Fig. \ref{Fig:pre_training_epoch}, Point-DAE with Transformer backbone converges faster than that with DGCNN. 
	In this work, we set the number of pre-training epochs as 300 in all experiments except those with DGCNN in Sec. \ref{Subsec:fine-tune}, where we pre-train the model for $1,200$ epochs.

	\begin{figure}[tb]
		\centering\includegraphics[width=.73\linewidth]{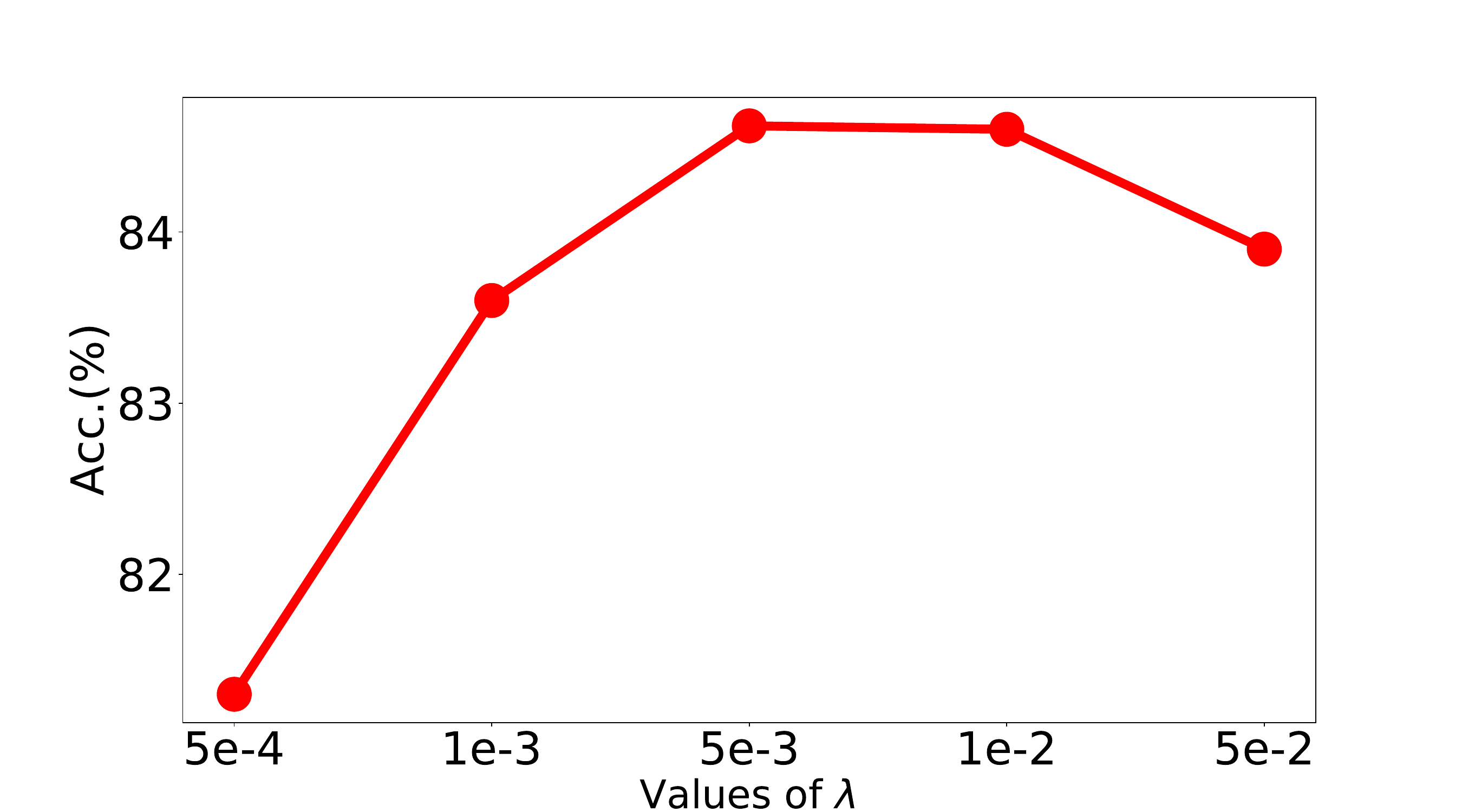}
		\caption{Results (\%) with different values of $\lambda$.
		}\label{Fig:lambda}
	\end{figure}
	
	\textbf{Hyper-parameter $\lambda$.}
	Results with different values of $\lambda$ are presented in Fig. \ref{Fig:lambda}.
	We empirically set $\lambda$ = 0.005 by default.

%
	
	\vspace{0.1cm}
	{\color{black}
	\textbf{Denoising Vs. Augmentation.} 
	Besides using corruptions in the denoising process as shown in Eq. (\ref{Equ:pretrain}), we introduce two degenerated versions by validating them as augmentation techniques, leading to the following two objectives:
	\begin{equation} \label{Equ:corruption_as_augmentation}
	\min_{\mathrm{E},\mathrm{D}} \mathbb{E}_{\X\in \mathbb{X}} \mathcal{L}\left(\mathrm{D}(\mathrm{E}(\mathrm{Mask}(\mathrm{Aff}(\X)))), \mathrm{Mask}(\mathrm{Aff}(\X)) \right),
	\end{equation}
	
	\begin{equation} \label{Equ:corruption_as_augmentation_affineonly}
	\min_{\mathrm{E},\mathrm{D}} \mathbb{E}_{\X\in \mathbb{X}} \mathcal{L}\left(\mathrm{D}(\mathrm{E}(\mathrm{Mask}(\mathrm{Aff}(\X)))), \mathrm{Aff}(\X) \right),
	\end{equation}
	where $\mathrm{Mask}(\cdot)$ and $\mathrm{Aff}(\cdot)$ represent the local masking and global affine transformation corruptions, respectively. 
	In Eq. (\ref{Equ:corruption_as_augmentation}), both masking and affine transformation are employed as augmentations in a vanilla auto-encoder framework, leading to the C1 variant in Tab. \ref{Tab:corruption_as_augmentation}. Masking is utilized for denoising, while affine transformation serves as the augmentation in Eq. (\ref{Equ:corruption_as_augmentation_affineonly}), corresponding to the C2 variant in Tab. \ref{Tab:corruption_as_augmentation}.
As shown in Tab. \ref{Tab:corruption_as_augmentation}, using corruptions for denoising significantly outperforms the use of them as augmentation, evidenced by the notable improvement of C3 over C1 and C2. 
These analyses validate the effectiveness of our proposed method.
}
	\begin{table}[h!]
		\centering
		\caption{\textcolor{black}{Classification results (\%) with different roles of corruption, where `Aug.' and `Den.' are abbreviations for augmentation and denoising, respectively.}
  } \label{Tab:corruption_as_augmentation}
		\begin{tabular}{l|cc|cc}
			\toprule
			&Affine & Masking       & DGCNN & Transformers \\
			\midrule
			C1 & Aug.   & Aug.             & 77.1    & 76.9  \\
			C2 & Aug.   & Den.            &  81.8    &  83.0 \\
                \midrule
			C3 & Den.   & Den.            & \textbf{85.5}    & \textbf{84.6}  \\
			\bottomrule
		\end{tabular}
	\end{table}

	\vspace{0.1cm}
	\textbf{Object Poses of Pre-training Data.}  
	We also reveal that object poses of pre-training data play important roles in self-supervised point cloud learning, which is less investigated in the community. 
	More specifically, samples are manually aligned to category-wise canonical poses in the popular pre-training datasets such as ShapeNet \cite{shapenet2015} and ModelNet \cite{wu20153d}.
	In other words, the canonical poses implicitly carry the manual category annotation, which is unintentionally utilized by the SSL algorithms, including the vanilla auto-encoder, masked modeling and Point-DAE, as shown in Fig. \ref{Fig:dataset_variants}.
	This does not conform to the practical scenarios where an unlabeled sample should not be manually assigned with any category-aware canonical pose.
	
	One intuitive solution to overcome the dataset limitation is assigning each point cloud with a random pose, leading to the `Random-Pose' setting. However, object pose plays an essential role in point cloud understanding \cite{zhao2022rotation,li2021closer} and it should not be neglected. To make appropriate use of the pose information, we promote to automatically estimate the canonical pose for each sample with the singular value decomposition (SVD) strategy \cite{golub1971singular,zhao2022rotation}, resulting in the `SVD-Pose' setting,
	where poses of samples in the same category are roughly aligned without any manual effort. In contrast, the vanilla dataset with manual poses is termed as `Manual-Pose'. 
	
	As illustrated in Fig. \ref{Fig:dataset_variants}, results with both `Manual-Pose' and `SVD-Pose' training data significantly outperform that with `Random-Pose' data, demonstrating the importance of object poses in the pre-training dataset. Although the `Manual-Pose' dataset has been widely used by the community and produces the best results, we recommend the `SVD-Pose' setting for practical consideration, where the object poses are properly used without any manual effort.
	Viewed from the perspective of sample corruptions, the effectiveness of masking and affine transformation corruptions still holds under all the three dataset settings. Additionally, in all the three settings, combining masking and affine transformation corruptions leads to improved results over individual corruption, validating their complementary.

	\vspace{0.1cm}
	\textbf{Visualization.} 
	As illustrated in Fig. \ref{Fig:reconstruct_vis}, the overall shape is reconstructed while the local details are hardly captured with the DGCNN backbone, due to the serious input corruptions of masking and affine transformation. 
	For results with the Transformer backbone, both the global shape and local patches are well constructed, thanks to the decomposition of the construction objective.
	
	{\color{black}
		\vspace{0.1cm}
	\textbf{Reconstruction Results and Downstream Performance.}
	We find that the reconstruction performance of a corruption type does not fully determine its performance in the downstream tasks. The reconstruction results with different input corruptions are illustrated in Fig. \ref{Fig:vis_masking_affine}. 
	We can see that it is more challenging to reconstruct the vanilla input from affine transformed samples than their masked counterparts. 
	Given a locally masked input, we are able to reconstruct a completed version of it with many details. However, given a globally affine transformed input, only the overall shape can be reconstructed without many local details. This discrepancy can be attributed to the fact that affine transformation introduces a larger shape deformation and hence leads to a more challenging reconstruction task, as revealed by the enlarged CD in Fig. \ref{Fig:mask_affine_cdloss}.
	However, the low reconstruction performance of a corruption type does not mean that it will lead to low performance in the downstream tasks. As shown in Tab. \ref{Tab:results_corruptions} and Tab. \ref{Tab:results_corruptions_transformer}, with the DGCNN backbone, the affine transformation outperforms the masking corruption on downstream tasks. While when the Transformer backbone is employed, masking corruption works better than affine transformation. In either case, jointly applying affine transformation and masking corruptions leads to the best results.
	
	
	\begin{figure}[htb]
		\centering\includegraphics[width=.85\linewidth]{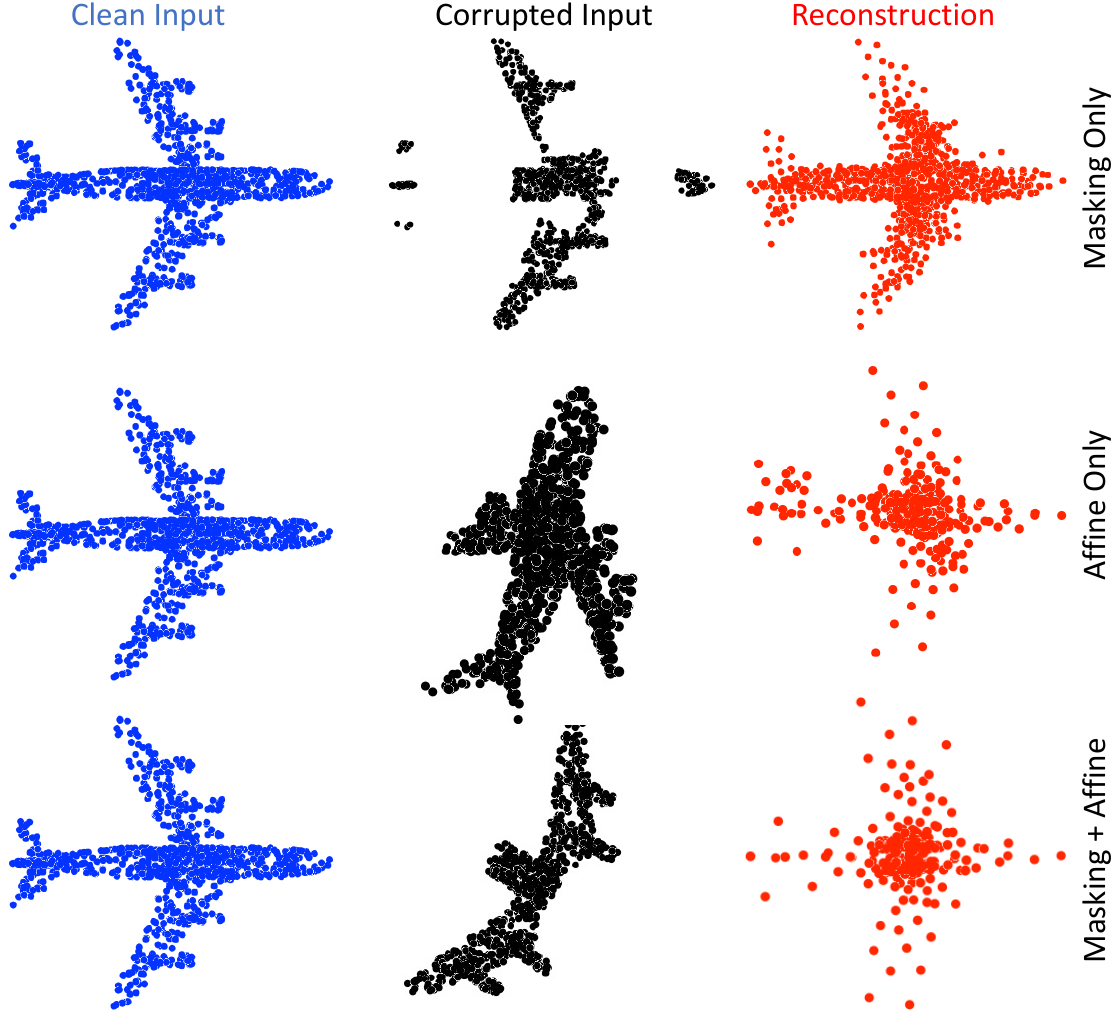}
		\vspace{-0.2cm}
		\caption{Reconstruction performance on DGCNN backbone with different input corruptions.} \label{Fig:vis_masking_affine}
		\vspace{-0.3cm}
	\end{figure} 
}
	
	\vspace{0.1cm}
	\textbf{Time complexity.} All Point-DAE variants with the same encoder backbone share almost the same training and test complexity, since they are only different in the used corruptions, which could be conducted within the dataloader in a parallel manner. 
	
	{\color{black}
		\vspace{0.1cm}
	\textbf{Discussions on the Effectiveness and Complementarity of Local and Global Corruptions.} 
	We found that local Masking and global Affine corruptions can respectively enhance the model's feature extraction capability of local and global structures. The effectiveness and complementarity of local and global structures in point cloud learning have been widely recognized in literature. For example, the graph attention block and spatial/channel-wise attention strategies are respectively proposed to exploit local and global structures for point cloud segmentation \cite{wang2019exploiting}. Similarly, a local spatial attention convolution and a global spatial attention module are proposed to respectively capture local structures and long-range contextual knowledge in \cite{du2022novel}. Unlike these methods that design specific modules to extract local and global features, in this paper we enhance the model's ability to extract information at different scales by introducing different corruptions in pre-training. Specifically, local Masking enhances the model's perception of local structures through reconstructing masked local regions, while global Affine enhances the model's perception of global shapes through restoring the deformed shapes. As illustrated in Fig. \ref{Fig:vis_masking_affine}, the model pre-trained with Masking corruption can better reconstruct local regions, whereas the model pre-trained with Affine corruption mainly reconstructs the global shape. 
	
	In addition to the visualization of reconstruction results, we further analyze the saliency maps with different pre-trained models in the shape recognition task. Specifically, we add a linear classification head to the pre-trained encoder. We fix the parameters of the pre-trained model and only train the newly added classification head on the ScanObjectNN dataset with the ObjBG setting. Then, we calculate the saliency maps by computing the gradient of the class score with respect to the input point cloud following \cite{simonyan2013deep}. The saliency maps are visualized in Fig. \ref{Fig:attention_vis}, where larger values represent higher importance to shape recognition. When using pre-trained encoders with the Affine corruption, we can see that points with higher attention scores tend to be evenly distributed across the global shape. In contrast, when using pre-trained encoders with Drop-Local corruption, points with higher attention scores are concentrated in local regions. This verifies that, compared to local Masking, global Affine corruption can enhance the model's perception capability of global shapes.

	\begin{figure}[t!]
		\centering\includegraphics[width=.9\linewidth]{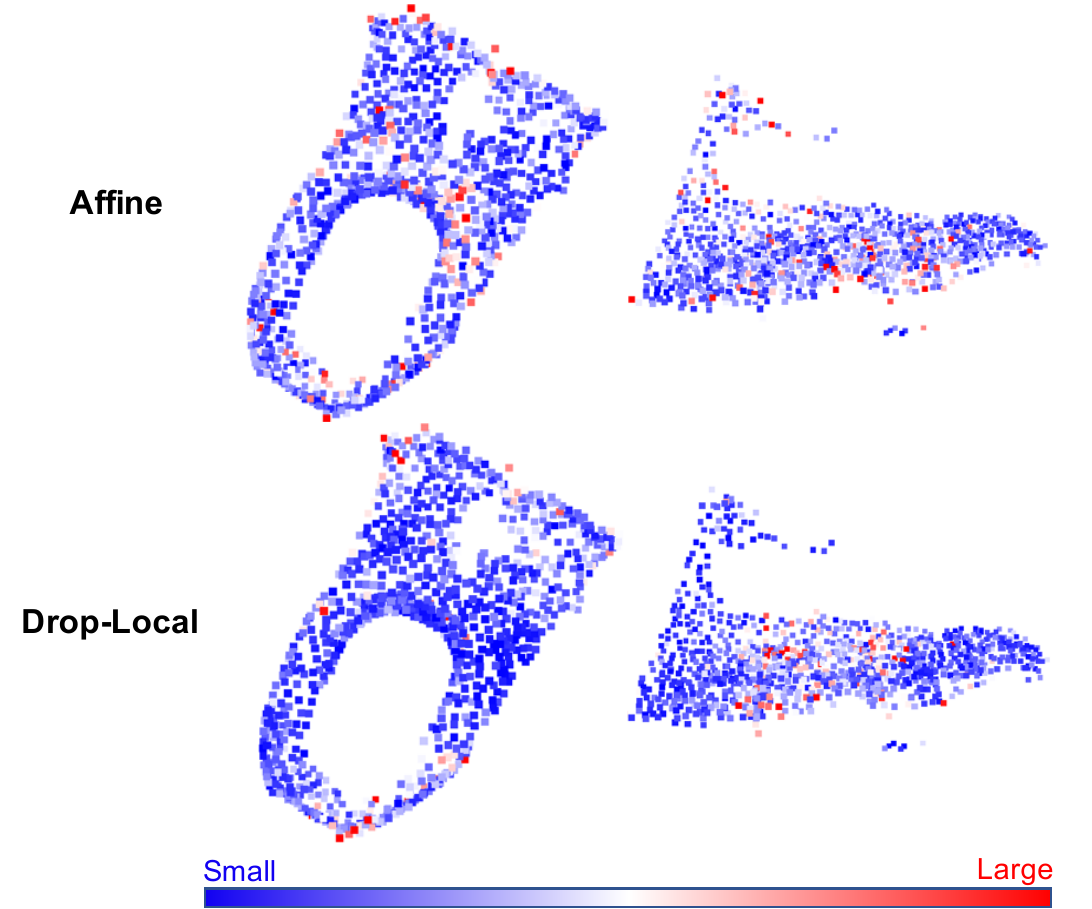}
		\caption{\textcolor{black}{Qualitative visualization of saliency maps on the ScanObjectNN dataset using pre-trained encoders, where blue and red colors represent small and large attention scores, respectively.}} \label{Fig:attention_vis}
  \vspace{-0.1cm}
	\end{figure}

	\begin{table}[t!]
		\centering
		\caption{\textcolor{black}{Quantitative analyses of attention map with various input corruptions on the ScanObjectNN dataset using pre-trained encoders. `CD' represents the Chamfer Distance between the top 20\% of points with high attention scores and all points of the whole shape. A smaller CD indicates that the pre-trained model focuses more on the global shape.  `Acc. (\%)' denotes the recognition accuracy under the linear probing protocol.}
  } \label{Tab:topcd_acc_corruptions}
		\begin{tabular}{l|cc|cc}
			\toprule
			& \multicolumn{2}{c|}{DGCNN} & \multicolumn{2}{c}{Transformer} \\
			\cmidrule{2-5}
			& CD & Acc. (\%)   & CD & Acc. (\%)  \\
			\midrule
			Affine             &  5.52    & 84.4  & 4.98 & 79.8  \\
			Local Masking &  6.71    & 80.8  & 5.61 & 82.7 \\
			Affine + Local Masking &  5.96   & 85.5  &  5.17 & 84.6 \\
			\bottomrule
		\end{tabular}
      \vspace{-0.1cm}
	\end{table}

	Besides qualitative visualization, we conduct quantitative analyses on the saliency maps with different corruptions. We select the top 20\% of points with high attention scores and calculate their Chamfer Distance (CD) with all points of the whole shape. A smaller CD indicates that the pre-trained model focuses more on the global shape. As shown in Tab. \ref{Tab:topcd_acc_corruptions}, compared to local Masking, the model pre-trained with Affine corruption achieves a significantly smaller CD, verifying its focus on the global shape. Combining global Affine and local Masking balances the extraction of local and global information, achieving a median CD but the best downstream recognition performance.

    It is also noticed that the effectiveness of Affine and Masking corruptions depends on the model architecture. For instance, Affine corruption is more effective with CNN-like models (\eg, PointNet++ \cite{qi2017pointnet++} and DGCNN \cite{wang2019dynamic}), while Masking corruption performs better with Transformer architectures, as presented in Tab. \ref{Tab:topcd_acc_corruptions}. This is because CNN-like architectures such as DGCNN and PointNet++ can effectively process and aggregate local information by design, inherently equipped with strong capabilities for local information extraction.  Therefore, using Affine can bring greater benefits to enhance the model's global perception. Conversely, Transformer-based methods are designed with a wider receptive field, so the benefits brought by global Affine corruption are less pronounced, but local Masking can provide greater benefits to Transformers in enhancing local information extraction. Furthermore, the patch-based processing in Transformers aligns better with patch-based local masking and reconstruction strategies. Regardless of the network structures, combining global Affine and local Masking corruptions brings further improvement, verifying the complementarity of these two corruptions.
	}

		\begin{table} [h!] 
		\centering
		\caption{\textcolor{black}{Linear probing results of Point-DAE on the ObjBG setting of ScanObjectNN dataset, where only the affine transformation corruption is applied.}
  } \label{Tab:T_net_ana}
    \vspace{-0.1cm}
		\begin{tabular}{l|cc}
			\toprule
			Backbones                               & \#Params & Acc. (\%) \\ 
			\midrule
			PointNet with T-Net                & 3.5M & 73.8 \\
			PointNet w/o T-Net (our adopted) & 0.8M  & \textbf{77.4} \\
			\bottomrule
		\end{tabular}
  \vspace{-0.3cm}
	\end{table}

		\vspace{0.1cm}
	{\color{black}
	\textbf{Failure Cases.}
As shown in Tab. \ref{Tab:T_net_ana}, the transformation module (\ie, T-Net) in PointNet \cite{qi2017pointnet} is less compatible with the affine transformation corruption.
This is mainly because the intrinsic transformation rectification mechanism in T-Net dilutes the influence of affine transformation. Therefore, we adopt the PointNet w/o T-Net as the default PointNet implementation. Though the adopted PointNet uses fewer parameters, it significantly outperforms other competitors, as shown in Sec. IV-C.

On the other hand, the severity of affine transformation significantly impacts the effectiveness of pre-training. Generally speaking, appropriately increasing the corruption level helps in learning effective representations. However, if the applied affine transformation is too strong, it can severely distort the input and hinder the extraction of features from normal object shapes. More detailed analyses on the magnitude of the 14 corruptions can be found in the Supplementary Material.
}

	\section{Conclusion}
	
	We comprehensively investigated the Point-DAE framework with $14$ corruption types. 
	Besides the popular masking corruption, we identified another effective corruption family, \ie, affine transformation, which disturbed all points globally and is complementary to the local masking strategy. 
	We validated the effectiveness of affine transformation corruption with various encoder backbones, including the traditional non-Transformer and the recent Transformer encoders. Especially, to alleviate the position leakage problem with the Transformer encoder, we decomposed the reconstruction of the complete point cloud into the reconstructions of detailed local patches and rough global shape, facilitating the pre-training efficacy. We validated the effectiveness of our method with various encoder backbones, under different evaluation protocols, and on extensive downstream point cloud tasks. 
	In future work, we will explore the generality of affine transformation corruption with more data modalities, \eg, 2D images and videos, under the denoising autoencoder framework.

	\bibliographystyle{IEEEtran_bst}
	\bibliography{IEEEabrv,egbib}

	\vfill
	
\end{document}